\documentclass{article}

\PassOptionsToPackage{dvipsnames,table}{xcolor}
\usepackage{xcolor}
\usepackage{PRIMEarxiv}
\fancypagestyle{firstpage}{
  \fancyhf{}
  
  \lfoot{\textit{Preprint}}
  \cfoot{\thepage}
}

\usepackage{amsmath,amssymb,amsfonts}
\usepackage{mathrsfs}
\usepackage{booktabs}
\usepackage{graphicx}
\usepackage{url}
\usepackage{multicol}
\usepackage{moreverb}
\usepackage{enumitem}
\usepackage{algorithm}
\usepackage{algorithmicx}
\usepackage{algpseudocode}
\usepackage{pifont}
\usepackage{xspace}
\usepackage{import}
\usepackage{tabularx}
\usepackage{listings}
\usepackage[numbers,square,comma,sort&compress]{natbib}
\usepackage[colorlinks,bookmarksopen,bookmarksnumbered,citecolor=red,urlcolor=red]{hyperref}

\newcommand{\ie}{\textit{i.e.}}
\newcommand{\eg}{\textit{e.g.}}

\newcommand{\nm}{GRAIL}

\newcommand{\NEUCRITIC}{V^{\mathrm{neu}}}
\newcommand{\LOGCRITIC}{V^{\mathrm{log}}}
\newcommand{\OP}[1]{\mathrm{#1}}

\newcommand\BibTeX{{\rmfamily B\kern-.05em \textsc{i\kern-.025em b}\kern-.08em
T\kern-.1667em\lower.7ex\hbox{E}\kern-.125emX}}

\usepackage{geometry}
\usepackage{textcomp}
\usepackage{listings}
\makeatletter

\newcommand\PrologPredicateStyle{}
\newcommand\PrologVarStyle{}
\newcommand\PrologAnonymVarStyle{}
\newcommand\PrologAtomStyle{}
\newcommand\PrologOtherStyle{}
\newcommand\PrologCommentStyle{}

\newif\ifpredicate@prolog@
\newif\ifwithinparens@prolog@

\lst@SaveOutputDef{`_}\underscore@prolog

\newcount\currentchar@prolog

\newcommand\@testChar@prolog%
{%
    \ifnum\lst@mode=\lst@Pmode%
        \detectTypeAndHighlight@prolog%
    \else
        \ifwithinparens@prolog@%
            \detectTypeAndHighlight@prolog%
        \fi
    \fi
    \global\predicate@prolog@false%
}

\newcommand\detectTypeAndHighlight@prolog
{%
    \def\lst@thestyle{\PrologAtomStyle}%
    \ifpredicate@prolog@%
    \def\lst@thestyle{\PrologPredicateStyle}%
    \else
    \expandafter\splitfirstchar@prolog\expandafter{\the\lst@token}%
    \expandafter\ifx\@testChar@prolog\underscore@prolog%
        \ifnum\lst@length=1%
            \let\lst@thestyle\PrologAnonymVarStyle%
        \else
            \let\lst@thestyle\PrologVarStyle%
        \fi
    \else
        \currentchar@prolog=65
        \loop
        \expandafter\ifnum\expandafter`\@testChar@prolog=\currentchar@prolog%
            \let\lst@thestyle\PrologVarStyle%
            \let\iterate\relax
        \fi
        \advance \currentchar@prolog by 1
        \unless\ifnum\currentchar@prolog>90
            \repeat
        \fi
    \fi
}
\newcommand\splitfirstchar@prolog{}
\def\splitfirstchar@prolog#1{\@splitfirstchar@prolog#1\relax}
\newcommand\@splitfirstchar@prolog{}
\def\@splitfirstchar@prolog#1#2\relax{\def\@testChar@prolog{#1}}

\def\beginlstdelim#1#2%
{%
    \def\endlstdelim{\PrologOtherStyle #2\egroup}%
    {\PrologOtherStyle #1}%
    \global\predicate@prolog@false%
    \withinparens@prolog@true%
    \bgroup\aftergroup\endlstdelim%
}

\newcommand\lang@prolog{Prolog-pretty}
\expandafter\lst@NormedDef\expandafter\normlang@prolog%
\expandafter{\lang@prolog}

\expandafter\expandafter\expandafter\lstdefinelanguage\expandafter%
{\lang@prolog}
{%
    language            = Prolog,
    keywords            = {},      % reset all preset keywords
    showstringspaces    = false,
    alsoletter          = (,
    alsoother           = @$,
    moredelim           = **[is][\beginlstdelim{(}{)}]{(}{)},
    MoreSelectCharTable =
    \lst@DefSaveDef{`(}\opparen@prolog{\global\predicate@prolog@true\opparen@prolog},
}

\newcommand\@ddedToOutput@prolog\relax
\lst@AddToHook{Output}{\@ddedToOutput@prolog}

\lst@AddToHook{PreInit}
{%
    \ifx\lst@language\normlang@prolog%
        \let\@ddedToOutput@prolog\@testChar@prolog%
    \fi
}

\lst@AddToHook{DeInit}{\renewcommand\@ddedToOutput@prolog{}}

\makeatother
\definecolor{PrologPredicate}{RGB}{65,105,225}%{000,031,255}
\definecolor{PrologVar}{RGB}{205,92,92}%{186,032,032}
\definecolor{PrologAnonymVar}{RGB}{000,127,000}
\definecolor{PrologAtom}     {RGB}{95,95,95}
\definecolor{PrologComment}  {RGB}{063,128,127}
\definecolor{PrologOther}    {RGB}{000,000,000}
\definecolor{backcolour}{rgb}{0.95,0.95,0.92}

\renewcommand\PrologPredicateStyle{\color{PrologPredicate}}
\renewcommand\PrologVarStyle{\color{PrologVar}}
\renewcommand\PrologAnonymVarStyle{\color{PrologAnonymVar}}
\renewcommand\PrologAtomStyle{\color{PrologAtom}}
\renewcommand\PrologCommentStyle{\itshape\color{PrologComment}}
\renewcommand\PrologOtherStyle{\color{PrologOther}}

\lstdefinestyle{Prolog-pygsty}
{
language     = Prolog-pretty,
upquote      = true,
stringstyle  = \PrologAtomStyle,
commentstyle = \PrologCommentStyle,
literate     =
    {:-}{{\PrologOtherStyle :-}}2
{,}{{\PrologOtherStyle ,}}1
{.}{{\PrologOtherStyle .}}1
}

\usepackage{amsmath,amsfonts,bm}

\def\eqref#1{equation~\ref{#1}}
\def\1{\bm{1}}

\DeclareMathAlphabet{\mathsfit}{\encodingdefault}{\sfdefault}{m}{sl}
\SetMathAlphabet{\mathsfit}{bold}{\encodingdefault}{\sfdefault}{bx}{n}

\newcommand{\E}{\mathbb{E}}

\definecolor{darkgreen}{RGB}{0,150,0}

\graphicspath{{assets/}}

\title{GRAIL: Autonomous Concept Grounding for Neuro-Symbolic Reinforcement Learning}

\author{
  Hikaru Shindo\thanks{Correspondence: \texttt{hikaru.shindo@tu-darmstadt.de}} \\
  Technical University of Darmstadt
  \And
  Henri R\"o\ss ler \\
  Technical University of Darmstadt
  \AND
  Quentin Delfosse \\
  Technical University of Darmstadt \\
  Google Intrinsic
  \And
  Kristian Kersting \\
  Technical University of Darmstadt \\
  hessian.AI \\
  German Research Center for Artificial Intelligence (DFKI) \\
  Centre for Cognitive Science, TU Darmstadt
}

\begin{document}
\maketitle
\thispagestyle{firstpage}

\begin{abstract}
    Neuro-symbolic Reinforcement Learning (NeSy-RL) combines symbolic reasoning with gradient-based optimization to achieve interpretable and generalizable policies. Relational concepts—such as ``left of'' or ``close by''—serve as foundational building blocks that structure how agents perceive and act. However, conventional approaches require human experts to manually define these concepts, limiting adaptability since concept semantics vary across environments. We propose GRAIL (Grounding Relational Agents through Interactive Learning), a framework that autonomously grounds relational concepts through environmental interaction. GRAIL leverages large language models (LLMs) to provide generic concept representations as weak supervision, then refines them to capture environment-specific semantics. This approach addresses both sparse reward signals and concept misalignment prevalent in underdetermined environments. Experiments on the Atari games \textit{Kangaroo}, \textit{Seaquest}, and \textit{Skiing} demonstrate that GRAIL matches or outperforms agents with manually crafted concepts in simplified settings, and reveals informative trade-offs between reward maximization and high-level goal completion in the full environment.
\end{abstract}

\section{Introduction}
\label{sec:introduction}

Deep reinforcement learning (RL) has achieved remarkable progress in recent years, driving advancements in critical fields such as autonomous driving and robotics. Deep neural networks, capable of learning policies across diverse tasks without prior domain knowledge~\citep{mnih2013atari,schulman2017proximal,badia2020agent57,Bhatt2024crossq}, have thus become the foundation of modern RL. Despite their success, these black-box models are prone to shortcut learning, exploiting action strategies that may be imperceptible to humans~\citep{liu2024role, delfosse2025deep}.
For instance, in Atari Pong, deep RL agents often gravitate toward behavior that focuses on the opponent's position instead of tracking the ball~\citep{delfosse2024interpretable}, demonstrating limited generalization when the environment is altered even slightly.

To overcome the limitations of neural approaches, RL has increasingly incorporated symbolic reasoning through logic-based policies~\citep{Jiang_2019_NLRL,kimura2021neuro,Cao22GALOIS,Delfosse_2023_NUDGE} and programmatic frameworks~\citep{sun2020program,verma18programmatically,Lyu19aaai,cappart2021combining,kohler2024interpretable}. These methodologies offer transparency, revisability, enhanced generalization, and facilitate curriculum learning. Nevertheless, they remain heavily dependent on human-provided inductive biases—requiring domain experts to hard-code essential concepts or logic rules—and often struggle to capture fine-grained, low-level behaviors. This reliance fundamentally constrains the flexibility and expressiveness of symbolic systems.

Research in philosophy and cognitive science has long maintained that human generalization capabilities stem from the ability to perceive the world through \textit{concepts}~\citep{Bruner_1956_Thinking,Rosch_1973_Natural_Categories}.
Concepts represent abstract attributes or relations common across sets of entities~\citep{Archer_1966_Definition_Concept}; for example, by color, shape, or positional relation to others. While concept learning has been explored in visual reasoning tasks such as Visual Question Answering~\citep{Mao_2019_NSCL,Hsu_2023_What_Is_Left,Mao_2025_NeSy_Concepts} and robotic manipulation~\citep{Silver_2023_Predicate_Invention}, grounding concepts in RL tasks remains relatively uncharted.
Current RL approaches bypass this by manually specifying grounding functions~\citep{Jiang_2019_NLRL,Vouros_2022_DERRL,Delfosse_2023_NUDGE,Shindo_2024_BlendRL}, a practice feasible in simple domains but impractical when facing greater complexity or relational structure involving multiple objects.

\begin{figure}[t]
    \centering
    \includegraphics[width=\linewidth]{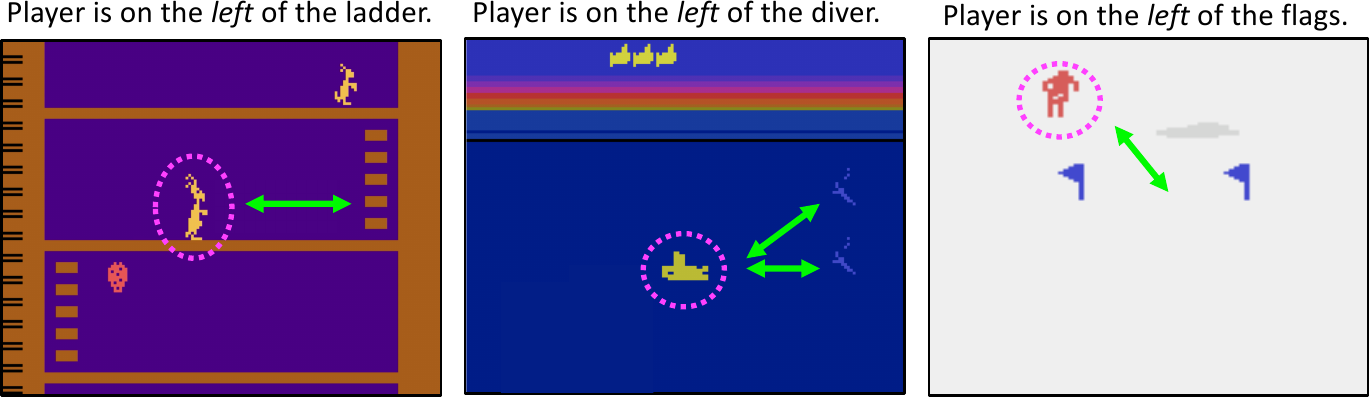}
    \caption{
        \textbf{Concept grounding is environment-dependent in neuro-symbolic reinforcement learning.} The spatial concept ``left of'' requires different interpretations across environments. In Kangaroo (left), the agent must verify that the player is both horizontally left of the ladder \emph{and} vertically aligned on the same platform. In Seaquest (middle), ``left of'' is defined more flexibly based on horizontal positioning regardless of vertical alignment. In Skiing (right), the relation is computed relative to the midpoint between flags rather than individual flag objects. While humans intuitively adapt these conceptual meanings to context, existing neuro-symbolic RL frameworks rely on manually hard-coded valuation functions for each environment~\citep{Shindo_2024_BlendRL}, severely limiting their scalability and adaptability to novel domains.}
    \label{fig:grail_intro}
\end{figure}

Figure~\ref{fig:grail_intro} illustrates this necessity across three Atari environments. Here, the agent must dynamically ground the “left of” relationship in different contexts: in Kangaroo (left), “left of ladder” entails being to the left of a ladder and on the same platform; in Seaquest (middle), “left of diver” simply means to the left of a diver irrespective of vertical alignment; in Skiing (right), “left of flags” requires identifying flags ahead and computing one's position relative to them. Currently, these varied conceptual groundings are hard-coded, which limits adaptability to new environments. This fundamental challenge raises an important research question: \emph{How can an agent learn to ground relational concepts autonomously through environment interactions?}

To address this, we propose GRAIL (Grounding Relational Agents through Interactive Learning), a novel framework that enables agents to ground relational concepts via experience. GRAIL builds upon BlendRL~\citep{Shindo_2024_BlendRL}, which employs a hybrid policy architecture—combining symbolic logic rules and differentiable neural networks—trained jointly with differentiable forward reasoning~\citep{Evans_2018_DILP,shindo23alphailp}. In BlendRL, each symbolic predicate is associated with a differentiable function that computes truth values over state observations, allowing the overall policy to seamlessly bind symbolic and neural reasoning.

GRAIL extends this by introducing a new method for concept grounding within the BlendRL framework, allowing each concept to be learned and adapted to its specific environment. Crucially, GRAIL leverages Large Language Models (LLMs) to provide general, high-level descriptions of concepts as weak supervision signals. For example, the LLM supplies a prototypical representation of “left,” which GRAIL then uses alongside environment feedback to train differentiable functions that maximize reward and align their outputs with the LLM-provided signal. This is accomplished by adding a novel loss term to the Proximal Policy Optimization (PPO)~\citep{schulman2017proximal} objective, encouraging alignment between learned concept groundings and guidance from LLMs.

\begin{figure}[t]
    \centering
    \includegraphics[width=\linewidth]{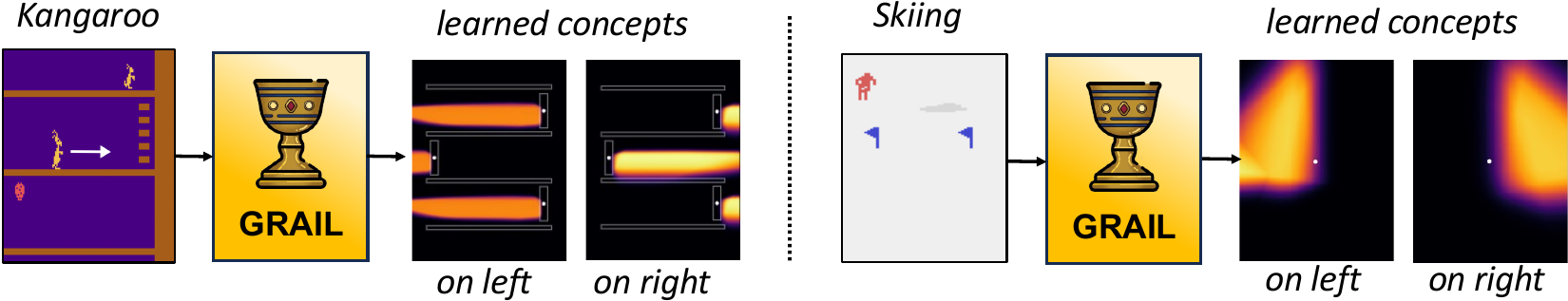}
    \caption{
        \textbf{GRAIL learns environment-specific concept groundings.}
        Given different environments, GRAIL autonomously discovers distinct interpretations of ``left'' and ``right.''
        In Kangaroo (left), the learned concepts activate along horizontal bands aligned with each platform, reflecting that ``left of ladder'' requires both horizontal offset and vertical alignment.
        In Skiing (right), activation extends diagonally above each flag, capturing the anticipatory nature of steering decisions during downhill movement.
        These heatmaps demonstrate that GRAIL adapts abstract relational concepts to the specific spatial structure of each environment.}
    \label{fig:grail_task_intro}
\end{figure}

Our experiments in the Atari environments Kangaroo, Seaquest, and Skiing demonstrate that GRAIL matches or outperforms both neural and neuro-symbolic baselines in a simplified setting, successfully discovering task-optimal concept groundings directly from interaction. As illustrated in Figure~\ref{fig:grail_task_intro}, GRAIL learns fundamentally different groundings of ``left'' and ``right'' depending on the environment—horizontal platform-aligned concepts in Kangaroo versus anticipatory diagonal concepts in Skiing.
In summary, our core contributions are:
\begin{itemize}
    \item We introduce GRAIL\footnote{Code is available at: https://github.com/ml-research/grail}, a framework that enables neuro-symbolic agents to ground relational concepts through environment interaction. GRAIL extends BlendRL by learning spatial relational concepts autonomously, using LLMs to provide high-level concept representations as weak supervision. The resulting policies are highly interpretable, expressed as first-order logic rules over the learned concepts.
    \item We formulate \emph{Concept Alignment} as a novel regularization term for PPO-based policy learning, encoding the degree to which the agent's learned concepts align with LLM-generated proxy functions. The resulting GRAIL learning framework navigates the inherent trade-off between reward maximization and semantically faithful concept grounding, and is among the first to discover spatial concept representations in this setting.
    \item We evaluate GRAIL on three challenging Atari environments: Kangaroo, Seaquest, and Skiing, which have not previously been tackled by neuro-symbolic agents without hard-coded relational concepts.
    We demonstrate that GRAIL matches or outperforms both the state-of-the-art neuro-symbolic baseline and a purely neural PPO baseline. Furthermore, we qualitatively analyze the relational concepts learned by GRAIL agents, showing that they acquire meaningful spatial groundings from experience without concept-level supervision.
\end{itemize}

\section{Background}
GRAIL builds on several foundational research areas, briefly reviewed in this section.
\subsection{Deep Reinforcement Learning}
\label{sec:DRL}

We model the environment as a Markov decision process (MDP), $\mathcal{M}=\langle\mathcal{S},\mathcal{A},P,R,\gamma\rangle$. The objective is to learn a policy $\pi_\theta(a_t \mid s_t)$ that maximizes the expected discounted return:
\begin{align}
    J(\theta) = \E_{\pi_\theta}\left[\sum_{t=0}^{T} \gamma^t r_t \right]
\end{align}
where $\gamma \in [0, 1]$ is the discount factor and $T$ is the episode length.

\subsubsection{Proximal Policy Optimization.}
\label{sec:ppo}
GRAIL optimizes policies using Proximal Policy Optimization (PPO)~\citep{schulman2017proximal} actor-critic method, that maintains both a policy (actor) $\pi_\theta$ and a value function (critic) $V_\phi$, evaluating the actor's decisions.
PPO estimates the advantage of each action using Generalized Advantage Estimation (GAE):
\begin{align}
    \label{eq:gae}
    \hat{A}_t^{\operatorname{GAE}(\gamma, \lambda)} = \sum_{i=0}^T (\gamma \lambda)^i \delta_{t+i}^{(1)}
\end{align}
where $\delta_t^{(1)} = r_t + \gamma V_\phi(s_{t+1}) - V_\phi(s_t)$ is the one-step TD residual and $\lambda$ controls the bias-variance trade-off. The policy is then updated by maximizing the following clipped surrogate objective:
\begin{align}
    \label{eq:ppo}
    L^{\operatorname{CLIP}}(\theta) = \E_{(s, a) \sim \pi_{\theta_{\operatorname{old}}}}\left[
        \min\left( r_t(\theta)\hat{A}_t, \operatorname{clip}\left( r_t(\theta), 1-\epsilon, 1+\epsilon \right) \hat{A}_t \right)
        \right]
\end{align}
where $r_t(\theta) = \pi_\theta(a_t \mid s_t)/\pi_{\theta_{\operatorname{old}}}(a_t \mid s_t)$ is the probability ratio between the current and previous policy, and $\epsilon$ is a clipping coefficient that constrains updates to a trust region, enabling stable reuse of experience data across multiple gradient steps.

\subsection{Neuro-Symbolic Reinforcement Learning}
\label{sec:NSRL}

GRAIL specifically builds upon the ideas of neuro-symbolic reasoning and learning with first-order logic.

\subsubsection{First-Order Logic.}
GRAIL uses \textit{First-Order Logic} (FOL) to encode world knowledge and actions in a logical and structured manner. A language in FOL, $\mathcal{L} = (\mathcal{P}, \mathcal{F}, \mathcal{D}, \mathcal{V})$, comprises predicate symbols $\mathcal{P}$, functors $\mathcal{F}$, constants $\mathcal{D}$ and variables $\mathcal{V}$.

An \textit{atom} $\mathtt{p(t_1, \dots, t_n)}$ is the smallest unit in a logical statement, where $\mathtt{t_1, \dots, t_n}$ are terms and $\mathtt{p}$ is a predicate of arity $\alpha(\mathtt{p}) = n$. Ground atoms (with only constant terms) have truth values. A \textit{Horn clause} takes the form $A \text{ :- } B_1, \dots, B_n$, where $A$ is the \textit{head} and $\{B_1, \dots, B_n\}$ is the \textit{body}, meaning if all body atoms are true, then $A$ must hold.

\subsubsection{Logic for Actions.}
GRAIL adopts first-order logic as the core language for representing both actions and states, enabling explicit reasoning throughout the agent's learning process with the logic programming framework~\citep{Lloyd84}. This perspective traces back to foundational work on logical reasoning about actions~\citep{Reiter2001action}; GRAIL follows and extends recent neuro-symbolic efforts such as~\citep{Delfosse_2023_NUDGE} by structuring policies with weighted first-order logic rules.

The predicate set $\mathcal{P}$ is split into \emph{action predicates} ($\mathcal{P}_A$) and \emph{state predicates} ($\mathcal{P}_S$). This separation empowers the agent to distinguish what it can \emph{do} from what it can \emph{know} about the world. The resulting \emph{Action-State Language} is defined by $(\mathcal{P}_A, \mathcal{P}_S, \mathcal{D}, \mathcal{V})$.
For illustration, consider the \textit{Kangaroo} environment (Figure~\ref{fig:grail_intro}), where action predicates may include $\mathcal{P}_A = \{\mathtt{go\_left}, \mathtt{go\_right}, \mathtt{jump}, \mathtt{idle}\}$, while state predicates could be $\mathcal{P}_S = \{\mathtt{left\_of}, \mathtt{closeby}, \ldots\}$. An \emph{action rule} takes the form $X_A \texttt{:-} X_S^{(1)}, \ldots, X_S^{(n)}$—the action is taken when all body conditions hold. For example, ``move right if left of a ladder'':
\begin{lstlisting}[
        mathescape,
        language=Prolog,
        style=Prolog-pygsty,
    ]
    go_right(O1):-type(O1,agent),type(O2,ladder),left_of(O1,O2).
\end{lstlisting}

\subsubsection{Differentiable Reasoning for RL.}
GRAIL is built upon \emph{differentiable logic programming}~\citep{Evans_2018_DILP,shindo2021dilpst,shindo23alphailp}, in which logical reasoning is realized through differentiable tensor operations, enabling end-to-end gradient-based optimization of symbolic representations.

\begin{figure}[t]
    \centering
    \includegraphics[width=\linewidth]{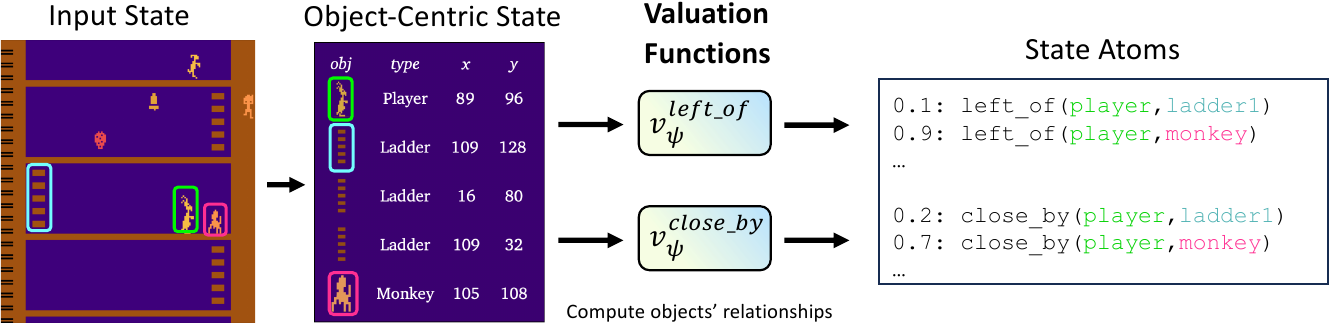}
    \caption{\textbf{Valuation functions evaluate relationships between objects.} These functions are typically hard-coded, limiting the applicability of neuro-symbolic reinforcement learning. GRAIL aims to learn these functions by aligning them with the correct concept directly from interactions with the environment.}
    \label{fig:grail_valuation}
\end{figure}

Figure~\ref{fig:grail_valuation} illustrates the computational flow. Raw states are first transformed into object-centric representations, where each object is described by its attributes (\eg, type, $x$- and $y$-coordinates). These representations are then processed by \emph{valuation functions}—differentiable parameterized functions that estimate the confidence of each state atom. For example, $v_\psi^{\mathtt{left\_of}}$ computes a soft confidence score for the $\mathtt{left\_of}$ predicate given a pair of objects. The outputs of these valuation functions form a weighted set of ground atoms that feed into the symbolic policy reasoning.

While valuation functions have traditionally been hand-crafted, limiting applicability and scalability, GRAIL provides a unified framework to \emph{learn} them directly from environment interaction while encouraging alignment with semantically meaningful concepts.
Because the entire reasoning pipeline is differentiable, GRAIL can optimize concept representations end-to-end via policy gradient methods, jointly improving task performance and concept quality.

\section{Related Work}
\label{sec:related_work}
GRAIL is built upon neuro-symbolic reinforcement learning, concept learning, and object-centric representation learning. We review each area and position our contributions accordingly.

\subsection{Neuro-Symbolic RL}
Relational Reinforcement Learning (Relational RL)~\citep{Dzeroski01RelationalRL, Kersting04Bellman, KerstingD08PolicyGradient, Lang12RelationalRLModel, DeepRelationalRLNeSy,Acharya_2023_Neuro_Symbolic_RL_Survey, golivand2025human} leverages logical representations and probabilistic reasoning to address RL challenges in structured, relational domains. The Neural Logic Reinforcement Learning (NLRL) framework~\citep{Jiang_2019_NLRL} is a pioneering effort to introduce Differentiable Inductive Logic Programming ($\partial$ILP)~\citep{Evans_2018_DILP} into RL. Here, $\partial$ILP facilitates the learning of generalized logic rules from examples using gradient-based optimization. NUDGE~\citep{Delfosse_2023_NUDGE} builds further by incorporating neurally-guided symbolic abstraction, drawing on significant progress in differentiable logic programming~\citep{shindo23alphailp,shindo2024neumann} to learn more complex programs. BlendRL~\citep{Shindo_2024_BlendRL} subsequently extends these ideas, combining symbolic and neural policies within a unified framework.

While these approaches demonstrate the effectiveness of learning logic-based policies, they share a common limitation: the reliance on \emph{manually} specified relational predicates, including explicit definitions of their semantics to compute confidence scores. As a result, adapting such methods to novel environments typically requires considerable manual effort to define suitable predicates and their underlying grounding functions. In contrast, GRAIL overcomes this bottleneck by automatically learning the \emph{grounding} of relational predicates---i.e., the valuation functions that define their semantics---directly from environment interaction, thus substantially broadening the applicability of neuro-symbolic RL.

\subsection{Concept Learning}
Learning \emph{concepts} is a fundamental challenge in artificial intelligence and machine learning.
Modeling concepts explicitly in the machine learning pipeline enhances interpretability and generalization of data-driven models~\citep{cbm,cbm_embedding,stammer2021cvpr_right,steinmann2025occb}.
Neuro-symbolic methods address this challenge by learning concepts from experience with symbolic programs~\citep{Mao_2025_NeSy_Concepts}, with an emphasis on complex visual reasoning with multiple objects and relations~\citep{Mao_2019_NSCL,sha2025naij,sha2025gestalt,wust2026synthesizing,deisam24shindo,vlol,Hsu_2023_What_Is_Left}.

However, these works focus on concept understanding in perception tasks or question answering; concept grounding in RL settings---\ie, learning \emph{what} relational predicates mean through environment interaction---remains underexplored.
GRAIL learns to ground relational concepts through interaction, guided by weak supervision from LLMs.

GRAIL draws on classical formalisms for describing concepts and actions abstractly.
Allen's interval algebra~\citep{AllenCalculus} provides a qualitative calculus over temporal intervals; the spatial predicates learned by GRAIL can be viewed as a continuous, learned counterpart of such qualitative relations.
Furthermore, action languages grounded in first-order and second-order logic have long been used to specify actions in planning and reinforcement learning~\citep{Reiter2001action}.
GRAIL follows the same tradition, representing policies as logic rules whose head atoms correspond to the actions executed by the agent.

\subsection{Object-Centric RL}
Object-centric decomposition is a fundamental pillar for achieving task generalization in reinforcement learning~\citep{delfosse2025deep}.
Object-centric reinforcement learning agents first need to transform unstructured state representations by decomposing visual inputs into object-centric states~\citep{locatello2020slotattention, SPACE2020, kipf2021SaVi, Delfosse2021MOC}.
These structured representations are increasingly integrated into RL pipelines, improving compositional generalization in model-free policies~\citep{haramatientity, chen2024slot, grandien2024interpretable} and enabling complex relational reasoning via object-level latent dynamics in model-based architectures~\citep{mosbach2025sold, dillies2025better, nishimoto2025object, bluml2025deep, feng2025learning}.
In the Atari domain, extracting such entity-level ground truth from raw pixels or RAM remains a fundamental challenge~\citep{OCAtari, luo2024end}.
To bridge this gap, recent approaches leverage pre-trained visual segmentations to construct sample-efficient spatial-temporal world models directly within these complex arcade environments~\citep{zhang2025objects, bluml2025deep}.

\section{GRAIL: Learning to Ground Relational Concepts}
GRAIL extends BlendRL~\citep{Shindo_2024_BlendRL} by replacing its hand-crafted valuation functions with \emph{learnable} differentiable grounding functions, guided by LLM-generated proxy concepts as weak supervision.
In the following, we outline the limitations of BlendRL's manual grounding and describe our proposed advances.

\begin{figure}
    \centering
    \includegraphics[width=\linewidth]{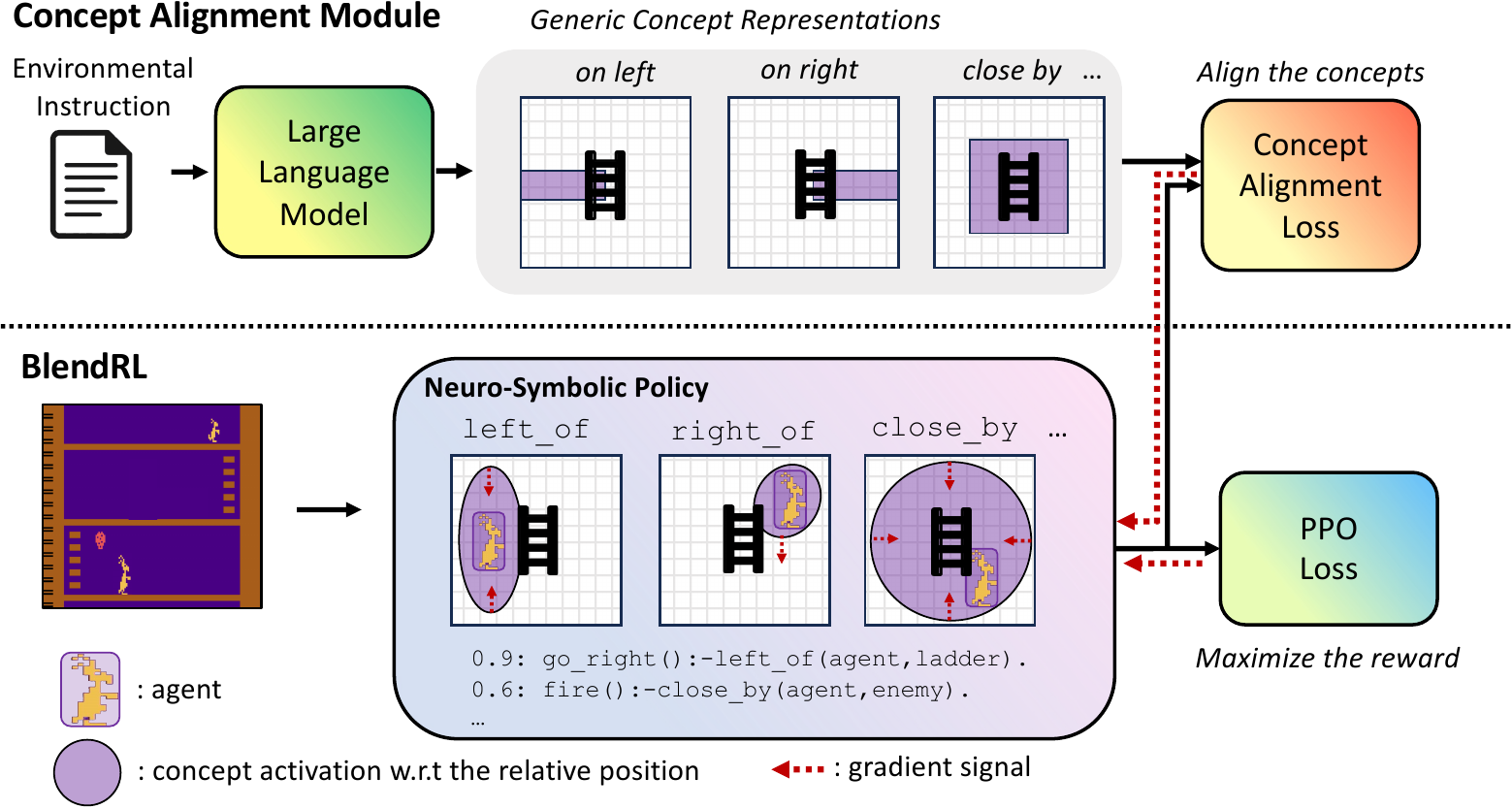}
    \caption{
        \textbf{GRAIL: Framework Overview.}
        GRAIL extends the BlendRL framework~\citep{Shindo_2024_BlendRL}, uniting neural and logic-based policies within a neuro-symbolic RL agent. Here, the logic policy consists of weighted rules expressed over predicates, where each predicate is equipped with a differentiable valuation function capturing abstract state relations. Unlike BlendRL, which relies on hand-crafted predicate valuations, GRAIL introduces \emph{concept alignment}: it leverages large language models (LLMs) to extract generic concept representations and incorporates a dedicated loss term that encourages the learned valuation functions to match these LLM-derived proxies. In this depiction, the spatial relations between agents and objects are visualized—purple areas indicate the normalized outputs of the corresponding valuation functions, \ie, the concept activations with respect to the relative positions of objects.
    }
    \label{fig:framework_overview}
\end{figure}
An overview of our approach is provided in Figure~\ref{fig:framework_overview}. The BlendRL framework trains neuro-symbolic policies to maximize expected reward using Proximal Policy Optimization (PPO)~\citep{schulman2017proximal}. In this setup, the logic policy is represented as a set of weighted rules over predicates, with each predicate associated with a differentiable valuation function to capture abstract state relations.

A central challenge arises in learning these spatial relations: aligning each predicate with its intended concept—the well-known \emph{symbol grounding problem}—is nontrivial. As a result, naively implementing predicates as neural networks and optimizing solely for reward frequently leads to poor or uninterpretable alignments.

To address both reward maximization and robust concept alignment, we introduce a concept grounding mechanism, depicted at the top of Figure~\ref{fig:framework_overview}. This module leverages a large language model (LLM) to extract general representations of relevant concepts, informed by environmental instructions that succinctly describe the task and domain. In essence, the LLM generates proxy representations for each concept (for example, specifying what ``left'' should look like in general). GRAIL then grounds these general, LLM-derived representations to the specifics of a given environment (such as Kangaroo), refining them through interaction and reward maximization.
While a generic ``left'' representation may not initially yield high performance, GRAIL improves this by learning to adapt and refine concept valuations according to environmental feedback. This allows the agent to achieve both generalization across tasks and strong environment-specific performance.
Crucially, this introduces an inherent trade-off: too strong an alignment signal constrains the agent to the LLM's generic priors and can impede reward maximization, while too weak a signal leaves the agent susceptible to degenerate or semantically meaningless groundings. We address this tension through an annealing schedule and a tunable alignment coefficient.

\subsection{The Hybrid Policy Reasoning and Learning}
\begin{figure}
    \centering
    \includegraphics[width=\linewidth]{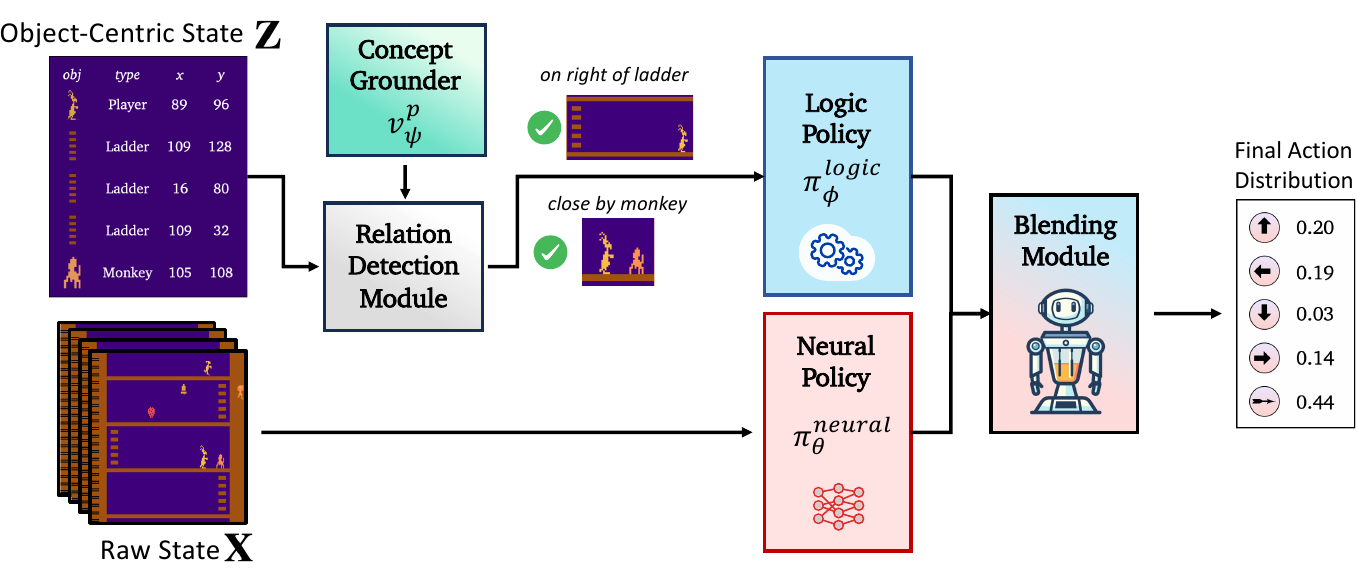}
    \caption{
        \textbf{GRAIL's Policy Reasoning.}
        A \textit{concept grounding module} takes object-centric features $\mathbf{Z}$ extracted from an image $\mathbf{X}$ and computes object relations via differentiable valuation functions $v_\psi^{\mathtt{p}}$.
        Those are then applied to a set of logical rules through forward reasoning to determine a \textit{logic policy}.
        Likewise, a \textit{blending module} utilizes the relational information to combine the logic policy with a \textit{neural policy} that operates on the sub-symbolic state.
        All components can be trained jointly.
    }
    \label{fig:grail_policy_reasoning}
\end{figure}

GRAIL inherits the hybrid policy architecture from BlendRL~\citep{Shindo_2024_BlendRL}, which combines neural and symbolic policies trained jointly.
We summarize this inherited architecture below for completeness.
The input state is represented by both a pixel-based and a symbolic representation, and the policy reasoning is depicted in Figure~\ref{fig:grail_policy_reasoning}.

\paragraph{Hybrid State Representations.}
GRAIL agents utilize two complementary forms of state representation: (i) \emph{pixel-based representations}, and (ii) \emph{object-centric representations}. The former comprise stacks of raw images directly provided by the environment and typically processed via convolutional neural networks~\citep{Mnih2015dqn}. The latter are extracted using object discovery models~\citep{Redmon_2016_YOLO, SPACE2020, Delfosse2021MOC, zhao2023fast} and consist of structured lists of objects with associated attributes (\eg, position, orientation, color), enabling explicit logical reasoning~\citep{zadaianchuk2021self, liu2021semantic, yoon2023investigation, wust2024pix2code, stammer2024neural}.
Alternatively, these states can be systematically extracted if supported by the environment. In the case of Atari, OCAtari~\citep{OCAtari} accomplishes this by reading the internal RAM state to produce structured object data.

Formally, the raw (sub-symbolic) state is denoted as $\mathbf{X} \in \mathbb{R}^{F \times W \times H \times C}$, representing the most recent $F$ frames of width $W$, height $H$, and $C$ channels. The symbolic (object-centric) state is denoted as $\mathbf{Z} \in \mathbb{R}^{n \times m}$, where $n$ is the number of detected objects and $m$ is the number of extracted properties per object.

\paragraph{Hybrid Policy Reasoning.}
Given both object-centric and pixel-based state representations, GRAIL conducts parallel neural and symbolic policy inference, and seamlessly combines their outputs through a blending mechanism.
This hybrid policy reasoning is composed of three main components:

\begin{enumerate}
    \item \textbf{Neural Policy:} $\pi^{\mathrm{neu}}: \mathbb{R}^{F \times W \times H \times C} \rightarrow [0, 1]^{A}$. This module is a neural network with parameters $\theta$, producing a probability distribution over actions from the pixel-based input $\mathbf{X}$. Standard implementations utilize convolutional neural networks~\citep{Mnih2015dqn, schulman2017proximal, hessel2018rainbow}, though visual transformers~\citep{chen2021decision, parisotto2020stabilizing} are equally compatible.
    \item \textbf{Logic Policy:} $\pi^{\mathrm{log}}: \mathbb{R}^{n \times m} \rightarrow [0, 1]^{A}$. Parameterized by $\phi$, this component is a differentiable forward reasoner~\citep{shindo23alphailp,shindo2024neumann} operating on object-centric representations (as visualized in Figure~\ref{fig:grail_policy_reasoning}). Policies are specified using FOL rules, where each rule comprises a head atom (the \emph{action}) and body atoms (the \emph{state predicates} serving as preconditions)~\citep{Reiter2001action, Delfosse_2023_NUDGE}.
    \item \textbf{Blending Module:} This component, parameterized by $\lambda$, is a differentiable function that computes a soft weighting between the neural and logic policies. The blender can be realized as either an explicit logic-based function ($B_\lambda: \mathbb{R}^{F \times n \times m} \rightarrow [0, 1]$), an implicit neural network ingester of pixel states ($B_\lambda: \mathbb{R}^{F \times W \times H \times C} \rightarrow [0, 1]$), or a hybrid of both. While logic-based blending is inherently interpretable, it presumes the presence of sufficient inductive biases—if these are absent, a neural blending approach may be preferable for adaptivity.
\end{enumerate}

The agent's final action distribution is obtained by blending the neural and logic policies:
\begin{align}
    \pi = \beta \cdot \pi^{\mathrm{neu}}(\mathbf{X}) + (1 - \beta) \cdot \pi^{\mathrm{log}}(\mathbf{Z}),
\end{align}
where $\beta = B_\lambda(\mathbf{Z}) \in [0, 1]$ denotes the blending weight inferred from the current symbolic (object-centric) state, $\pi^{\mathrm{neu}}$ is parameterized by $\theta$, and $\pi^{\mathrm{log}}$ by $\phi$.
All modules of the agent are optimized jointly using PPO.

To compute value estimates, separate critics process the respective state modalities: the \emph{neural critic}, $\NEUCRITIC: \mathbb{R}^{W \times H \times C} \rightarrow \mathbb{R}$, for sub-symbolic states, and the \emph{logic critic}, $\LOGCRITIC: \mathbb{R}^{E \times D} \rightarrow \mathbb{R}$, for symbolic states. These are then blended analogously:
\begin{align}
    V = \beta \cdot \NEUCRITIC(\mathbf{X}) + (1 - \beta) \cdot \LOGCRITIC(\mathbf{Z}).
\end{align}

\paragraph{Policy Optimization.}
Policy optimization in our framework builds upon the standard PPO objective (Eq.~\ref{eq:ppo}), which comprises loss terms for the value function, clipped policy ratio, and action entropy regularization. To further encourage the agent to leverage both neural and logic policies, we adopt and extend the BlendRL regularization for blending:
\begin{align}
    \label{eq:blender_entropy}
    H(B_\lambda) = -\beta \cdot \log \beta - (1 - \beta) \cdot \log (1 - \beta)
\end{align}
This \textit{blender entropy} quantifies the uncertainty or diversity in the blending coefficient $\beta$, which softly allocates control between the neural ($\beta$) and logic ($1-\beta$) policies. By encouraging higher entropy, the agent is discouraged from fully collapsing onto either policy and is instead incentivized to employ them both as appropriate for the state.

The final BlendRL loss function, which we denote $L^{\operatorname{BlendRL}}$, thus takes the following form:
\begin{align}
    \label{eq:blendrl_loss}
    L^{\operatorname{BlendRL}} = \mathbb{E}\Big[
        c_{\operatorname{VF}} \cdot L^{\operatorname{VF}}
        - L^{\operatorname{CLIP}}
        - c_{\operatorname{AE}} \cdot H(\pi) - c_{\operatorname{BE}} \cdot H(B_\lambda)
        \Big]
\end{align}
where $L^{\operatorname{VF}}$ is the mean-squared value function error, $L^{\operatorname{CLIP}}$ is the clipped surrogate objective (Eq.~\ref{eq:ppo}), $H(\pi)$ is the action entropy, and $H(B_\lambda)$ is the blender entropy defined above. The coefficients $c_{\operatorname{VF}}$, $c_{\operatorname{AE}}$, and $c_{\operatorname{BE}}$ weight the respective terms. All components operate on the hybrid policy $\pi$ and value function $V$ as defined above.

\paragraph{The Concept-Grounding Bottleneck.}
Up to this point, we have presented the hybrid policy reasoning approach, which enables agents to reason abstractly and perform reactive decision-making. However, a central limitation of this framework is its dependence on user-supplied concept grounding, specifically the requirement for \emph{hand-crafted valuation functions} to define predicates such as “left.” This reliance restricts the framework’s applicability across different environments, since concepts like “left” can have varying semantics depending on context, as illustrated in Figure~\ref{fig:grail_intro}.

To overcome this challenge, we introduce a \textit{concept grounding module} that leverages large language models (LLMs) to automatically generate proxy functions for each extensional predicate. LLMs offer generic, intuitive representations of concepts, serving as a form of conceptual prior knowledge about how these predicates are commonly understood. By incorporating LLM-generated proxies, GRAIL augments the BlendRL framework with an additional source of supervision—referred to as \textit{concept alignment}—that guides the learning of environment-specific grounding for abstract concepts.

\subsection{Grounding Spatial Concepts in Environments}
\label{sec:grounding}

Learning to ground abstract concepts within specific environments is a crucial capability of our neuro-symbolic architecture. \textit{Concept grounding} refers to the process by which abstract, symbolic predicates—such as $\mathtt{left\_of}$ or $\mathtt{close\_by}$—are mapped to context-dependent, observable, object-centric features obtained from the environment. In \nm, this is achieved through the learning of differentiable \textit{valuation functions} that output soft truth values for each predicate by processing the relational configuration of detected objects.

Rather than relying on static, hand-crafted rules, we employ parameterized and differentiable functions to evaluate spatial relational predicates. A simplistic method might use a shallow MLP that consumes the absolute positions of objects as input, but this generally fails to capture important invariances and generalization capabilities required in diverse environments. Three critical desiderata guide our improved design:
\begin{enumerate}
    \item \textbf{Translation Invariance:} Spatial relationships should not be affected by the simultaneous translation of all involved objects. Thus, we use relative coordinates, such as differences $(x_1 - x_j,\, y_1 - y_j)$, instead of absolute positions.
    \item \textbf{Normalization:} We normalize these coordinate differences by the width and height of the scene, ensuring all offset vectors $\bigl(\frac{x_1-x_j}{W},\, \frac{y_1-y_j}{H}\bigr)$ are scaled to $[-1, 1]^2$. This supports robustness to varying scene sizes.
    \item \textbf{Generality:} Although the logic programs in Figure~\ref{fig:logic_programs} employ only binary spatial relations, our framework is designed to handle predicates of arbitrary arity.
\end{enumerate}

Accordingly, we replace the hand-crafted spatial valuation functions in BlendRL with differentiable, parameterized valuation functions $v_\psi^{\mathtt{p}}: [-1,1]^2 \rightarrow [0,1]$, implemented as neural networks and trained jointly with the rest of the architecture. Given a binary spatial predicate $\mathtt{p}$ relating a reference object (object~1, typically the player) at position $(x_1, y_1)$ to a second object at position $(x_2, y_2)$, the valuation function takes normalized relative coordinates as input:
\begin{align}
    v_\psi^{\mathtt{p}}\!\left( \frac{x_1 - x_2}{W},\, \frac{y_1 - y_2}{H} \right) \in [0, 1],
\end{align}
where $W$ and $H$ denote the width and height of the scene, respectively. The normalized relative coordinates $\bigl(\frac{x_1 - x_2}{W},\, \frac{y_1 - y_2}{H}\bigr) \in [-1, 1]^2$ ensure translation invariance and robustness to varying scene sizes. While all spatial predicates in our experiments are binary, this formulation naturally extends to $n$-ary predicates by concatenating the normalized offsets for each additional object.

By optimizing the PPO-based BlendRL loss (Eq.~\ref{eq:blendrl_loss}) with respect to the parameters $\psi$, the agent is able to maximize reward by flexibly adapting its concept representations—effectively grounding abstract predicates to the specific spatial and contextual nuances of each environment.

\begin{figure}[t]
    \centering
    \textbf{(a) Policy Programs}\par\vspace{2pt}
    \begin{lstlisting}[
        mathescape, language=Prolog, style=Prolog-pygsty,
        aboveskip=2pt, belowskip=4pt,
    ]
up_ladder(X)    :- on_ladder(P,L), same_level_ladder(P,L).
right_ladder(X) :- left_of_ladder(P,L), same_level_ladder(P,L).
left_ladder(X)  :- right_of_ladder(P,L), same_level_ladder(P,L).
up_air(X)         :- oxygen_low(B).
up_rescue(X)      :- full_divers(X).
left_to_diver(X)  :- right_of_diver(P,D),
                      visible_diver(D), not_full_divers(X).
right_to_diver(X) :- left_of_diver(P,D),
                      visible_diver(D), not_full_divers(X).
up_to_diver(X)    :- deeper_than_diver(P,D),
                      visible_diver(D), not_full_divers(X).
down_to_diver(X)  :- higher_than_diver(P,D),
                      visible_diver(D), not_full_divers(X).
left_to_flag(X)  :-right_of_flag(P,F), right_oriented(P).
right_to_flag(X) :-left_of_flag(P,F), left_oriented(P).
noop(X)          :-right_of_flag(P,F), left_of_flag(P,R), straight_oriented(P).
\end{lstlisting}
    \textbf{(b) Blending Programs}\par\vspace{2pt}
    \begin{lstlisting}[
        mathescape, language=Prolog, style=Prolog-pygsty,
        aboveskip=2pt, belowskip=2pt,
    ]
neural_agent(X) :- close_by_monkey(P,M).
neural_agent(X) :- close_by_throwncoconut(P,TC).
logic_agent(X)  :- nothing_around(X).
neural_agent(X) :- close_by_enemy(P,E).
neural_agent(X) :- close_by_missile(P,M).
logic_agent(X)  :- visible_diver(D).
logic_agent(X)  :- oxygen_low(B).
logic_agent(X)  :- full_divers(X).
logic_agent(X)  :- true(X).
\end{lstlisting}
    \caption{Logic programs used by \textbf{(a)}~the logic actor and \textbf{(b)}~the blending module in three Atari environments, generated by LLMs following~\citep{Shindo_2024_BlendRL}. Unlike previous studies, where spatial predicates are hand-crafted by human experts, GRAIL grounds these predicates as parameterized differentiable functions. In Skiing, the blending program always delegates to the logic agent.}
    \label{fig:logic_programs}
\end{figure}
\subsection{Aligning Concepts with Semantic Priors}
\label{sec:concept_aligner}
While agents can learn to ground spatial concepts through trainable mechanisms, this alone does not guarantee that the resulting representations capture their correct semantic intent. For instance, the agent may confuse ``left'' with ``right,'' as there is nothing intrinsic in the learning process to prevent these concepts from being systematically swapped. This ongoing difficulty illustrates the classic symbolic grounding problem.

To overcome this limitation, we introduce the \textit{concept aligner} as an essential component of our framework. \textit{Concept alignment} refers to refining the agent’s learned, environment-specific concepts so they align with external semantic priors or generic conceptual knowledge—such as proxy functions derived from large language models (LLMs). By utilizing such weak supervision, the concept aligner encourages the learned valuation functions to faithfully represent the intended meanings of each concept.

Specifically, we employ LLMs to extract generic knowledge about spatial relations (e.g., how ``left'' should be interpreted in an abstract sense) and use this information to guide the alignment of learned valuation functions. Introducing this additional supervisory signal helps ensure that the agent’s internal representations are better aligned with universal, human-interpretable semantics. While integrating humans in the loop can provide high-quality, interpretable feedback~\citep{Stammer_2022_Interactive_Disentanglement,Natarajan_2025_HIL_AIIL}, it is often costly—especially when agents learn continually from interactive experiences. Leveraging LLM-generated supervision thus significantly reduces the effort needed to obtain meaningful feedback.

\begin{figure}
    \centering
    \includegraphics[width=\linewidth]{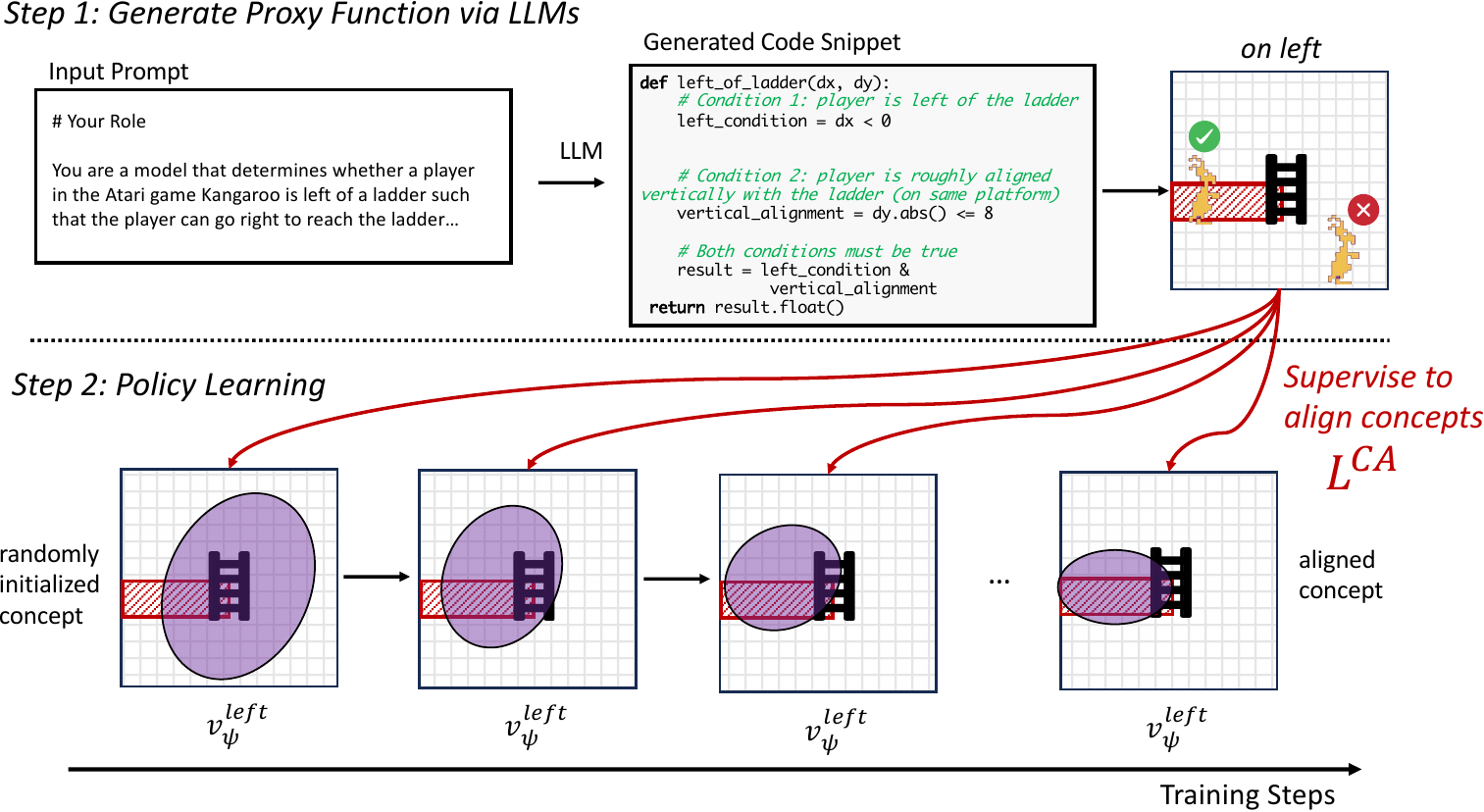}
    \caption{
        \textbf{GRAIL maximizes reward and aligns concepts semantically.}
        \textbf{Step 1:} We generate proxy functions for each spatial predicate using LLMs. These functions are represented as normalized activation maps, typically produced as code snippets and visualized over the 2D state space.
        \textbf{Step 2:} The proxy functions are incorporated as an additional supervision signal during policy optimization. This auxiliary signal guides the agent to semantically align its learned valuation functions with human-intended concepts---for example, preventing systematic confusion between ``left'' and ``right''.
        To achieve this, we introduce a binary cross-entropy loss $L^{\OP{CA}}$ that explicitly encourages the learned valuation functions to match their corresponding proxy functions.
    }
    \label{fig:concept_aligner}
\end{figure}
Figure~\ref{fig:concept_aligner} illustrates the overall concept aligner module.
We begin by leveraging large language models (LLMs) to generate generic, environment-agnostic representations of spatial concepts—so-called \emph{proxy functions}—by prompting the models with detailed descriptions of the task, relevant environmental features, and objectives.
Concretely, the LLM produces executable Python code that implements each spatial predicate $\mathtt{p}$ as a proxy function $g^{\mathtt{p}}: [-1,1]^2 \rightarrow [0,1]$, mapping a 2D relative offset to a soft truth value. The logic programs that define the policy structure (Figure~\ref{fig:logic_programs}) are also generated by LLMs following~\citet{Shindo_2024_BlendRL}.
These resulting proxy functions act as semantic priors, providing abstract ``templates'' of the intended meanings for each spatial relation.
Throughout the reinforcement learning process, these proxy functions are used as an auxiliary supervision signal: as the agent optimizes its actions for reward, the learned valuation functions are concurrently encouraged to align with the proxy functions.
This coupling helps ensure that the agent’s internal concept representations remain faithful to human-understandable semantics.
For example, if the agent’s policies incorrectly conflate the notions of ``left'' and ``right,'' the proxy functions will provide a corrective influence and steer the learned concepts toward the intended interpretation.
It is important to emphasize, however, that using proxy functions alone results in suboptimal performance, since they are only generic and not adapted to the specific environment.
Therefore, the process of \emph{grounding}—adapting concepts to their environment—is essential.
The alignment signal introduced by our approach substantially enhances this grounding by combining generic knowledge with environment-specific experience.

We now detail how the concept aligner incorporates proxy functions into our framework.
The core idea is to periodically compare the agent's learned valuation functions against the LLM-generated proxy functions over a dense grid of spatial offsets, and penalize any disagreement. Intuitively, this grid acts as a shared ``canvas'' on which both functions paint their activation maps; the concept alignment loss then measures how closely these two maps match for each predicate.

Concretely, we construct a $K \times K$ grid of offset vectors evenly distributed within the range $[-1, 1]^2$. For row $r \in \{1, \ldots, K\}$ and column $c \in \{1, \ldots, K\}$:
\begin{align}
    \Delta x_{r,c} = \frac{2c}{K+1} - 1, \qquad \Delta y_{r,c} = \frac{2r}{K+1} - 1.
\end{align}
At each training iteration, we evaluate both the learned valuation function $v_{r,c} = v_\psi^{\mathtt{p}}(\Delta x_{r,c}, \Delta y_{r,c})$ and the corresponding proxy function $\hat{v}_{r,c} = g^{\mathtt{p}}(\Delta x_{r,c}, \Delta y_{r,c})$ at every grid point.
The discrepancy between the learned concepts and the semantic priors is measured using the mean binary cross-entropy loss:
\begin{align}
    \label{eq:concept_alignment_loss}
    L^{\operatorname{CA}} =
    \frac{1}{|\mathcal{P}_e|}\sum_{\mathtt{p} \in \mathcal{P}_e} \frac{1}{K^2} \sum_{r,c}
    \operatorname{BCE}\!\left(\hat{v}_{r,c},\, v_{r,c}\right)
\end{align}
where $\mathcal{P}_e \subset \mathcal{P}$ denotes the set of extensional predicates whose semantics are to be aligned.
We choose binary cross-entropy (BCE) because both the learned valuation functions (sigmoid output) and the proxy functions produce values in $[0, 1]$ that can be interpreted as soft truth values.
BCE directly penalizes pointwise deviations in these truth values, which is appropriate when the proxy provides a reasonable \emph{shape} of the activation map.
Alternative objectives, such as ranking losses (which preserve only relative orderings) or contrastive losses (which encourage separation between positive and negative regions), may be more robust when proxy magnitudes are unreliable.

To integrate this semantic supervision into learning, we augment the original BlendRL objective (see Eq.~\ref{eq:blendrl_loss}) with our \textit{concept alignment loss} $L^{\operatorname{CA}}$, yielding the following total objective:
\begin{align}
    \label{eq:objective}
    L = L^{\operatorname{BlendRL}}
    + \left( 1 - \gamma_{\operatorname{CA}} \cdot \frac{t}{T} \right) \cdot c_{\operatorname{CA}} \cdot L^{\operatorname{CA}}
\end{align}
Here, $c_{\operatorname{CA}} \in \mathbb{R}_{\geq 0}$ is the \textit{concept alignment coefficient}, controlling the strength of the semantic prior, and $t \in \mathbb{N}$ ($0 \leq t \leq T$) is the current optimization step out of $T$ total steps.
The term $\gamma_{\operatorname{CA}} \in [0, 1]$ is a scheduling hyperparameter that determines the rate at which the influence of $L^{\operatorname{CA}}$ diminishes over training.
A value of $\gamma_{\operatorname{CA}} = 1$ leads the alignment loss to be annealed to zero by the end of training, while $\gamma_{\operatorname{CA}} = 0$ keeps it constant throughout.
This gradual attenuation reflects the role of the concept aligner: to provide helpful guidance during the early, ambiguous phase of training, but to allow final concept grounding to be informed primarily by environment-specific experience.
We examine the impact of varying $c_{\operatorname{CA}}$ and $\gamma_{\operatorname{CA}}$ in our ablation studies in our experiments.

\paragraph{Relation between grounding and alignment.}
\textit{Concept grounding} enables agents to learn what a concept means in a given environment, while \textit{concept alignment} ensures that this learned meaning remains semantically faithful to its general, language-level interpretation. By balancing these two objectives, GRAIL produces policies that are both reward-maximizing and interpretable.

\section{Experiments}
\label{sec:experiments}
We empirically assess our framework on a variety of Atari environments, focusing on both quantitative performance and the interpretability of learned spatial concepts. Our experimental study is structured to address the following research questions:

\begin{enumerate}
    \item[\textbf{Q1:}] Can GRAIL learn concept groundings that match the performance of hand-crafted valuation functions?
    \item[\textbf{Q2:}] Does GRAIL learn interpretable, environment-specific spatial concepts rather than simply replicating LLM proxies?
    \item[\textbf{Q3:}] Do GRAIL's learned concepts transfer to the full neuro-symbolic setting, and how do they affect the trade-off between reward maximization and goal completion?
    \item[\textbf{Q4:}] What failure modes arise in learned concept grounding, and where does concept misalignment persist?
\end{enumerate}

\subsection{Experimental Setup}
\label{sec:experimental_setup}
We compare GRAIL against two primary baselines: a neural baseline and a neuro-symbolic baseline.

\paragraph{Baselines.}
As the \textbf{neural baseline}, we use a CNN-based PPO agent~\citep{schulman2017proximal} with three convolutional layers (kernel sizes 8, 4, 3; strides 4, 2, 1), followed by a shared 512-dimensional fully connected layer for both the actor (18 actions) and critic (scalar value estimate).
As the \textbf{neuro-symbolic baseline}, we use BlendRL~\citep{Shindo_2024_BlendRL}, which has been shown to outperform prior neuro-symbolic RL methods such as NUDGE~\citep{Delfosse_2023_NUDGE} and NLRL~\citep{Jiang_2019_NLRL}. Since GRAIL builds upon BlendRL by replacing its hand-crafted valuation functions with learned ones, this comparison directly isolates the effect of our concept grounding mechanism.
In Stage~1, where the neural policy is disabled ($\beta=0$), BlendRL reduces to a purely logic-based policy akin to NUDGE; however, NUDGE was evaluated only on simpler environments and does not support learned valuation functions, precluding a direct comparison.
For GRAIL, we generate proxy functions using two LLMs—Claude4-Sonnet~\citep{Anthropic_2025_claude4sonnet} and GPT-4o~\citep{Openai_2025_chatgpt4o}—yielding two GRAIL variants.

\paragraph{Environments.}
We evaluate on three Atari environments from the Arcade Learning Environment (ALE)~\citep{bellemare13arcade}: Kangaroo, Seaquest, and Skiing.
Each environment demands different spatial concepts—platform-relative navigation in Kangaroo, underwater pursuit and rescue in Seaquest, and anticipatory steering in Skiing—providing complementary coverage of the challenges GRAIL addresses.
Prior neuro-symbolic RL methods such as NLRL~\citep{Jiang_2019_NLRL} and NUDGE~\citep{Delfosse_2023_NUDGE} were evaluated on simpler or synthetic environments; Atari games pose a substantially harder test due to high-dimensional visual input, dynamic multi-object scenes, and sparse rewards.
We use OCAtari~\citep{OCAtari} to extract object-centric features, representing each state in both pixel-based and object-centric modalities.

\paragraph{Metrics.}
We report the \emph{average episodic return} for quantitative comparison and the \emph{average goals achieved per episode} to measure high-level task completion.
We further provide qualitative analysis by visualizing the learned spatial concepts as heatmaps, allowing direct inspection of how GRAIL grounds relational predicates in each environment.

\paragraph{Training Protocol.}
We adopt a two-stage training protocol.
BlendRL~\citep{Shindo_2024_BlendRL} first trains end-to-end in a single stage. This is only possible because its valuation functions are hand-crafted, effectively bypassing the concept learning problem entirely.
Since GRAIL must \emph{learn} these functions, end-to-end training would require the agent to simultaneously learn two interdependent components: (1)~the meaning of each concept via valuation functions, and (2)~the importance of each logic rule whose predicates rely on those very concepts.
This creates a circular dependency: the agent cannot determine which rules are useful without knowing what the predicates that compose these rules mean, yet the predicates receive gradient signal only through the rules.
By first isolating concept learning in a simplified setting (Stage~1), we break this dependency and allow the valuation functions to converge to interpretable groundings before the full neuro-symbolic pipeline is trained (Stage~2).

\textbf{Stage 1: Logic Policy Training on Simplified Environment.}
We train only the logic policy and its valuation functions, disabling the neural policy ($\beta=0$) and removing all enemies, using HackAtari~\citep{Delfosse_2024_hackatari} tasks modifications.
Rule weights in the symbolic policy remain fixed. Rewards are restricted to high-level achievements (\eg, reaching the child in Kangaroo or rescuing six divers in Seaquest). Episodes are capped at 3000 steps with updates every 4 steps, for a total of 10 million steps.

\textbf{Stage 2: Joint Neuro-Symbolic Training on Complete Environment.}
We freeze the learned valuation functions and train the neural policy and blending module \emph{from scratch} in the full environment. The neural policy and blending weights are randomly initialized; only the spatial concept groundings are carried over from Stage~1. Enemies are reactivated, there is no episode length restriction, and updates use a step size of 1. The reward structure awards 20 points for level completion and 1 point for each other reward. This stage runs for 60 million steps.

\paragraph{Optimization Details.}
All parameters are optimized using PPO with respect to the joint objective (Eq.~\ref{eq:objective}). Each iteration samples 128 steps from the current policy across parallel environments. Advantages are estimated via GAE (Eq.~\ref{eq:gae}) with $\gamma = 0.99$ and $\lambda = 0.95$.

The loss coefficients are $c_{\operatorname{VF}} = 0.5$, $c_{\operatorname{AE}} = 0.01$, and $c_{\operatorname{BE}} = 0.01$, with clipping parameter $\epsilon = 0.1$. We use Adam with a linearly decayed learning rate from $2.5 \times 10^{-4}$ and gradient clipping at $0.5$. Parameters are updated for 10 epochs per iteration with 32 parallel environments.
We sweep over $c_{\mathrm{CA}} \in \{0.03,\, 0.1,\, 0.3,\, 1.0\}$ and report results for the best-performing setting.

\section{Results}
\label{sec:results}

We now present the empirical results for both training stages individually.

\subsection{Performance comparison on Atari environments}
\label{sec:exp_stage1}

\begin{table}[t]
    \centering
    \begin{tabularx}{\textwidth}{l *{3}{>{\centering\arraybackslash}X}}
        \toprule
        \textbf{Model}          & \textbf{Kangaroo} & \textbf{Seaquest} & \textbf{Skiing} \\
        \midrule
        NeuralPPO               & $1045_{\pm 577}$       & $453_{\pm 93}$                & $-5492_{\pm 2496}$       \\
        BlendRL (no CA)         & $1280_{\pm 372}$       & $953_{\pm 14}$                & $\circ\; -5086_{\pm 128}$        \\
        BlendRL+Expert          & $\mathbf{3683}_{\pm 29}$ & $\mathbf{983}_{\pm 93}$        & ---                      \\
        \midrule
        BlendRL+GPT-4o         & $1112_{\pm 40}$         & $783_{\pm 63}$              & $-5171_{\pm 124}$         \\
        BlendRL+Claude         & $910_{\pm 63}$          & $559_{\pm 46}$              & $-5388_{\pm 60}$                  \\
        \midrule
        \textbf{GRAIL (GPT-4o)} & $3540_{\pm 131}$       & $874_{\pm 40}$                & $-5253_{\pm 95}$         \\
        \textbf{GRAIL (Claude)} & $\circ\; 3625_{\pm 36}$ & $\circ\; 981_{\pm 162}$     & $\mathbf{-5021}_{\pm 22}$ \\
        \bottomrule
    \end{tabularx}
    \caption{
        \textbf{Average episodic return on the simplified environment.}
        GRAIL consistently matches or outperforms all baselines, including BlendRL with expert-written valuations.
        \textbf{Bold} = best; $\circ$ = second best. Higher is better ($\uparrow$) for Kangaroo and Seaquest; less negative is better ($\uparrow$) for Skiing.
        Averages over 3 seeds (100 episodes); standard deviations in subscript.
        BlendRL+Expert results are from the original paper~\citep{Shindo_2024_BlendRL}; BlendRL+LLM variants use fixed proxy functions without learned grounding.
        ``No CA'' denotes learned valuation functions without concept alignment ($c_{\OP{CA}} = 0$).
        Concept-alignment coefficients: $c_{\OP{CA}} = 0.3$ (Kangaroo), $c_{\OP{CA}} = 0.3$/$0.1$ (Seaquest, GPT-4o/Claude), $c_{\OP{CA}} = 0.1$ (Skiing).
    }
    \label{tab:results_return}
\end{table}

To address \textbf{Q1}, we evaluate agents on the Atari environments Kangaroo, Seaquest, and Skiing. Table~\ref{tab:results_return} reports the average episodic returns during the initial training phase, in which only the logic policy is active and the neural module is disabled. Both GRAIL and BlendRL with hand-crafted valuation functions achieve high scores, significantly outperforming the purely neural agent. This demonstrates that GRAIL can effectively ground spatial concepts and attain performance on par with policies designed using expert knowledge.

In contrast, except for Skiing, BlendRL variants that directly employ LLM-generated proxy functions---from GPT-4o or Claude---perform substantially worse. This result underscores the limitation of using LLM outputs as direct replacements for expert-designed functions, and highlights the strength of GRAIL's concept alignment strategy: rather than adopting LLM-generated functions verbatim, GRAIL treats them as supervision signals, enabling it to adapt its spatial semantics to the structure of each environment.

\begin{figure}[t]
    \centering
    \includegraphics[width=\linewidth]{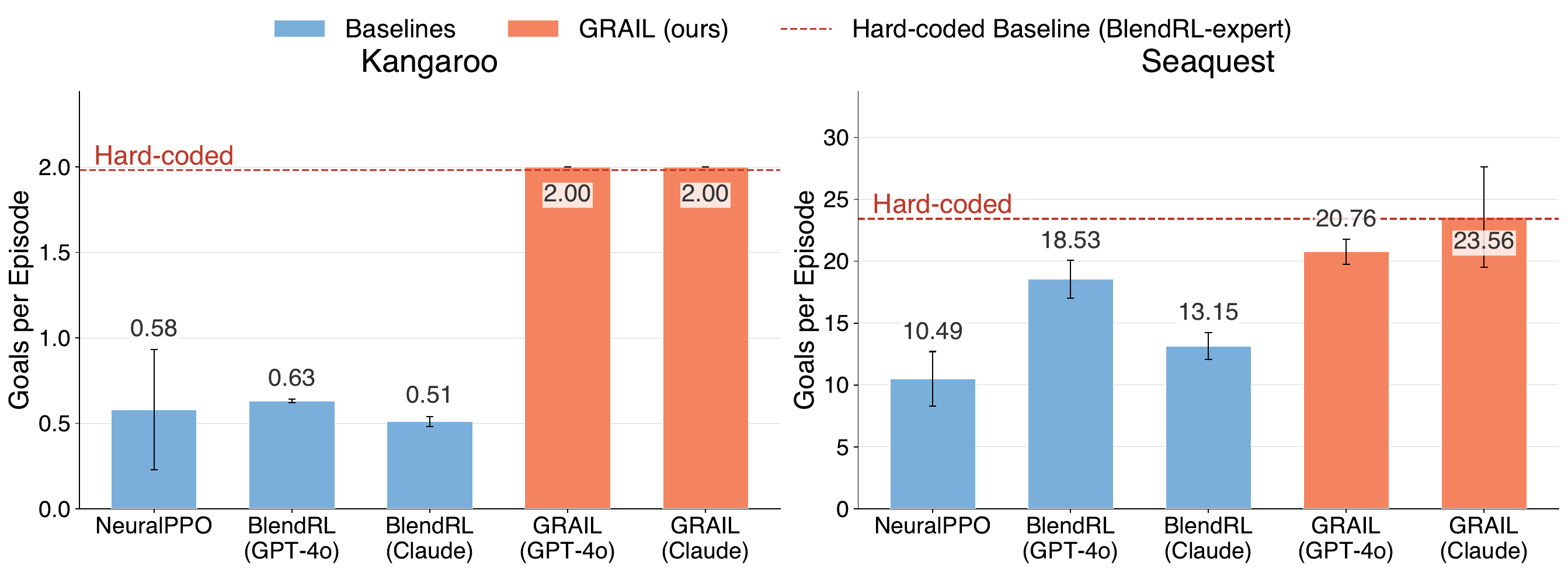}
    \caption{
        \textbf{GRAIL achieves high-level goals reliably by grounding spatial concepts.}
        Average number of goals achieved per episode on the simplified environment (Stage~1, enemies removed, logic policy only).
        In Kangaroo, a goal corresponds to reaching the child at the top platform; in Seaquest, a goal corresponds to successfully rescuing all six divers and surfacing.
        The red dashed line indicates the hard-coded BlendRL-expert baseline.
        Both GRAIL variants (GPT-4o and Claude) match or exceed this baseline, achieving 2.00 goals per episode in Kangaroo and over 20 in Seaquest.
        In contrast, purely neural agents and agents using raw LLM proxy functions as direct valuations fall significantly short---particularly BlendRL (Claude) in Seaquest (13.15), highlighting the insufficiency of unrefined LLM-generated concepts.
        Averages are computed over 3 seeds (100 episodes each); error bars indicate standard deviation.
        }
    \label{fig:results_goals}
\end{figure}

We further compare approaches by the number of high-level goals achieved per episode.
In Kangaroo, the goal is to reach the top of the screen; in Seaquest, to rescue all six divers while managing a depleting oxygen level. Figure~\ref{fig:results_goals} presents the average goals achieved per episode.
Both GRAIL and BlendRL with hand-crafted valuation functions exhibit similarly high success rates and clearly outperform the purely neural agent, indicating that GRAIL reliably completes tasks without converging on suboptimal strategies.
In these sparse-reward settings, purely neural agents tend to gravitate toward locally rewarding but ultimately ineffective behaviors---such as repeatedly firing at enemies for minor points rather than pursuing the main objectives. By leveraging LLMs to guide neuro-symbolic policies without being constrained by fixed proxy functions, GRAIL overcomes these limitations and achieves reliably goal-directed behavior even in the absence of hand-designed valuation functions.

\subsection{Interpretability of Learned Spatial Concepts}
\begin{figure}
    \centering
    \includegraphics[width=\linewidth]{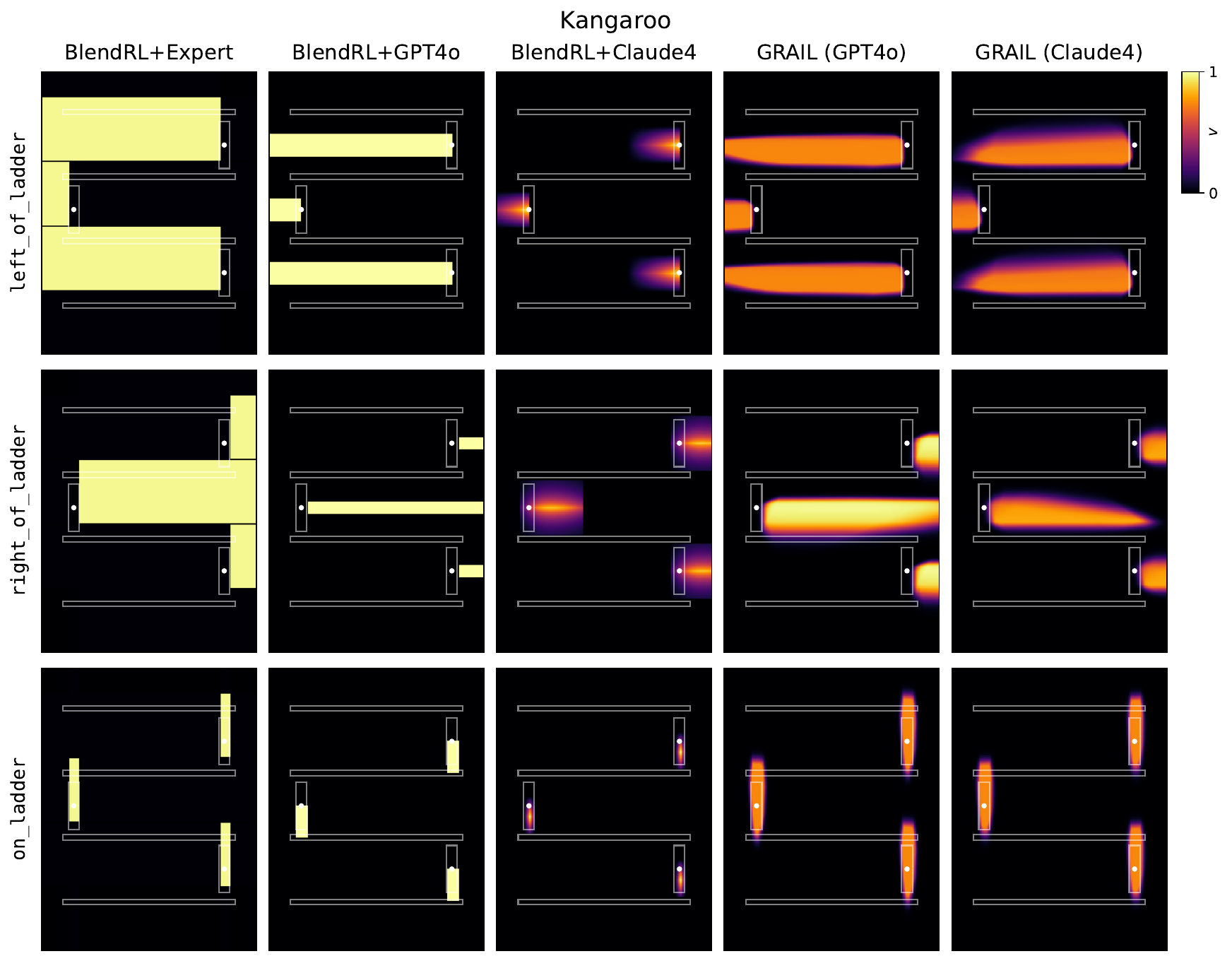}
    \caption{
        \textbf{GRAIL produces interpretable spatial concepts that capture subtle environmental details (Kangaroo).}
        Heatmaps indicate the truth values produced by the valuation functions as the player's position varies within the scene.
        Each white dot denotes the location of a ladder, relative to which truth values for the spatial predicates are evaluated.
        The results are visualized (from left to right) for: hand-crafted valuation functions (BlendRL+Expert), proxy functions generated by GPT-4o and Claude, and the valuation functions learned by GRAIL under weak supervision from either proxy.
        Platform and ladder outlines are depicted as white boxes.
    }
    \label{fig:kangaroo_1st}
\end{figure}

To address \textbf{Q2}, we examine the interpretability of the learned spatial concepts by comparing GRAIL’s valuation functions to both the hand-crafted functions from BlendRL and the LLM-generated proxy functions in the Kangaroo environment (Figure~\ref{fig:kangaroo_1st}).

Although both Claude and GPT-4o capture the general semantics of spatial concepts, they fail to ground them accurately within the game’s layout.
For instance, GPT-4o assigns high truth values for $\mathtt{left\_of\_ladder}$ and $\mathtt{right\_of\_ladder}$ across nearly the full width of each floor but with overly narrow vertical extent, while Claude’s proxy covers nearly the entire height but sharply truncates activation based on horizontal distance from the ladder.
These discrepancies reveal a fundamental limitation: without environment-specific adaptation, LLM-generated proxy functions lack the precision required to serve as direct valuations of spatial predicates.

GRAIL addresses this gap by treating the proxy functions as flexible supervision rather than fixed definitions, allowing the agent to refine concept representations through environmental feedback and the reward signal.
As a result, the learned spatial concepts are precisely tailored to Kangaroo’s layout---correcting the shortcomings of either proxy---and the resulting logic policy matches both the performance and interpretability of BlendRL’s hand-crafted functions while significantly surpassing either LLM alone.
This confirms that GRAIL does not merely replicate proxy functions but learns spatial concepts that are well-aligned with the environment’s structure.
We note that the heatmaps shown correspond to the best-performing seed; qualitatively similar spatial patterns emerge across all seeds despite minor variations in activation boundaries, as reflected in the standard deviations reported in Table~\ref{tab:results_return}.

\begin{figure}
    \centering
    \includegraphics[width=0.8\linewidth]{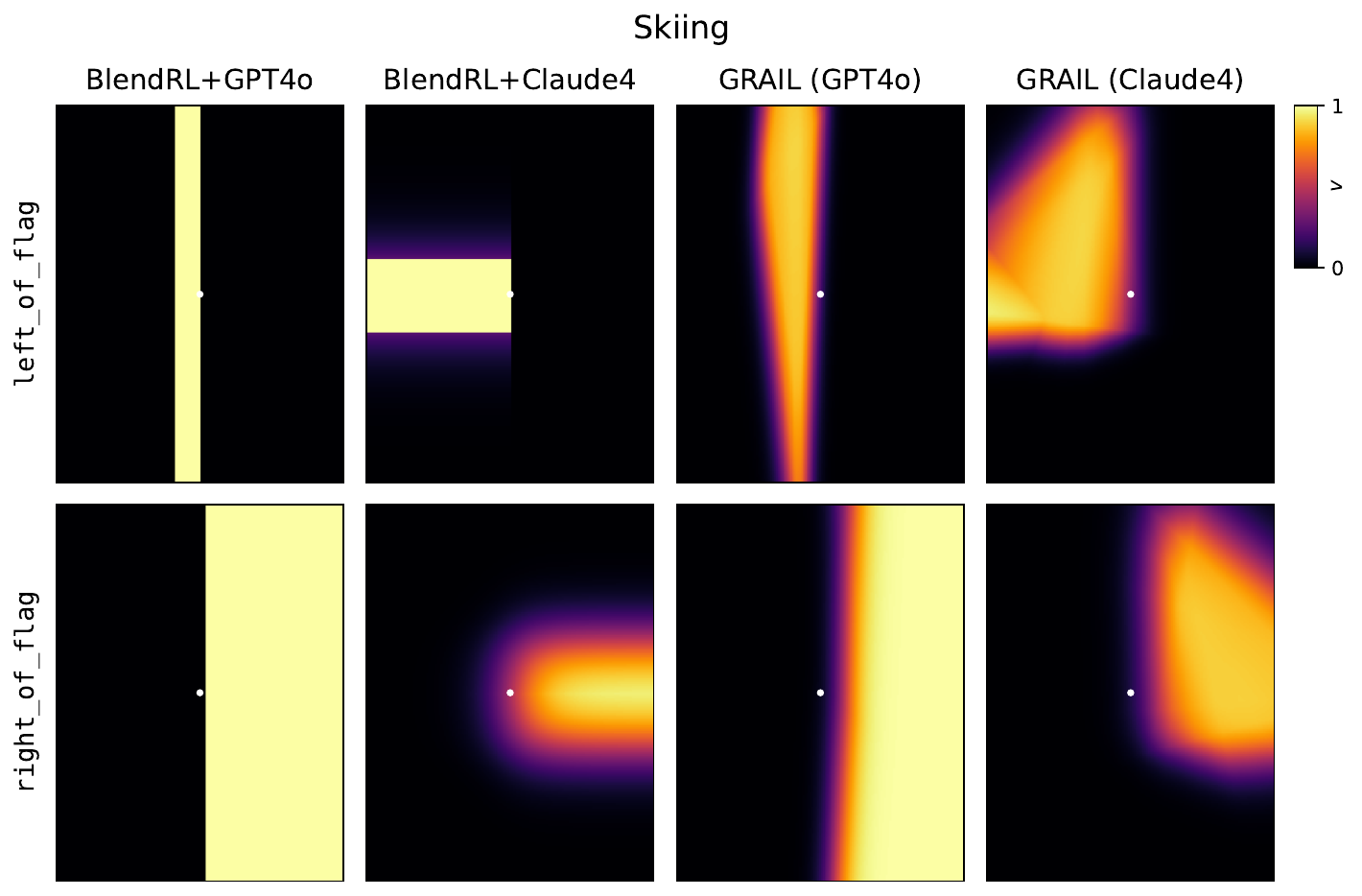}
    \caption{
        \textbf{Learned spatial concepts in Skiing.}
        Heatmaps show truth values for $\mathtt{left\_of\_flag}$ and $\mathtt{right\_of\_flag}$ as the player's position varies.
        Each white dot marks the position of a flag gate.
        Without concept alignment (BlendRL+GPT-4o, BlendRL+Claude), the proxy functions produce overly simplistic patterns—e.g., a narrow vertical strip or broad horizontal bands.
        In contrast, GRAIL learns asymmetric, environment-adapted concepts that concentrate activation in the relevant diagonal regions ahead of the player, reflecting the downhill direction of movement in Skiing.
    }
    \label{fig:skiing_1st}
\end{figure}

We observe a similar pattern in Skiing (Figure~\ref{fig:skiing_1st}), where the agent must learn the concepts $\mathtt{left\_of\_flag}$ and $\mathtt{right\_of\_flag}$ to navigate between flag gates.
The LLM-generated proxy functions fail to capture the environment-specific semantics: GPT-4o produces a narrow vertical strip for $\mathtt{left\_of\_flag}$, while Claude generates broad horizontal bands—neither accounts for the vertical structure of the task.
In Skiing, the player moves downhill and must identify whether it is left or right of an upcoming flag \emph{before} reaching it. This requires the learned concepts to incorporate a vertical margin: activation should extend above the flag's position, reflecting the anticipatory nature of the steering decision.
GRAIL with Claude successfully captures this skiing-specific semantics. The learned $\mathtt{left\_of\_flag}$ and $\mathtt{right\_of\_flag}$ heatmaps show activation concentrated above and to the relevant side of each flag, demonstrating that the agent has discovered that ``left of a flag'' in Skiing means being to the left \emph{and} slightly ahead of it.
This result highlights the adaptability of GRAIL: starting from generic LLM priors that only encode a na\"ive notion of horizontal direction, the framework autonomously learns environment-specific concept groundings that account for the vertical dynamics of the task.

\subsection{Concept Learning in Joint Neuro-Symbolic Policy Training}
\label{sc:exp_stage2}

\begin{figure}
    \centering
    \includegraphics[width=\linewidth]{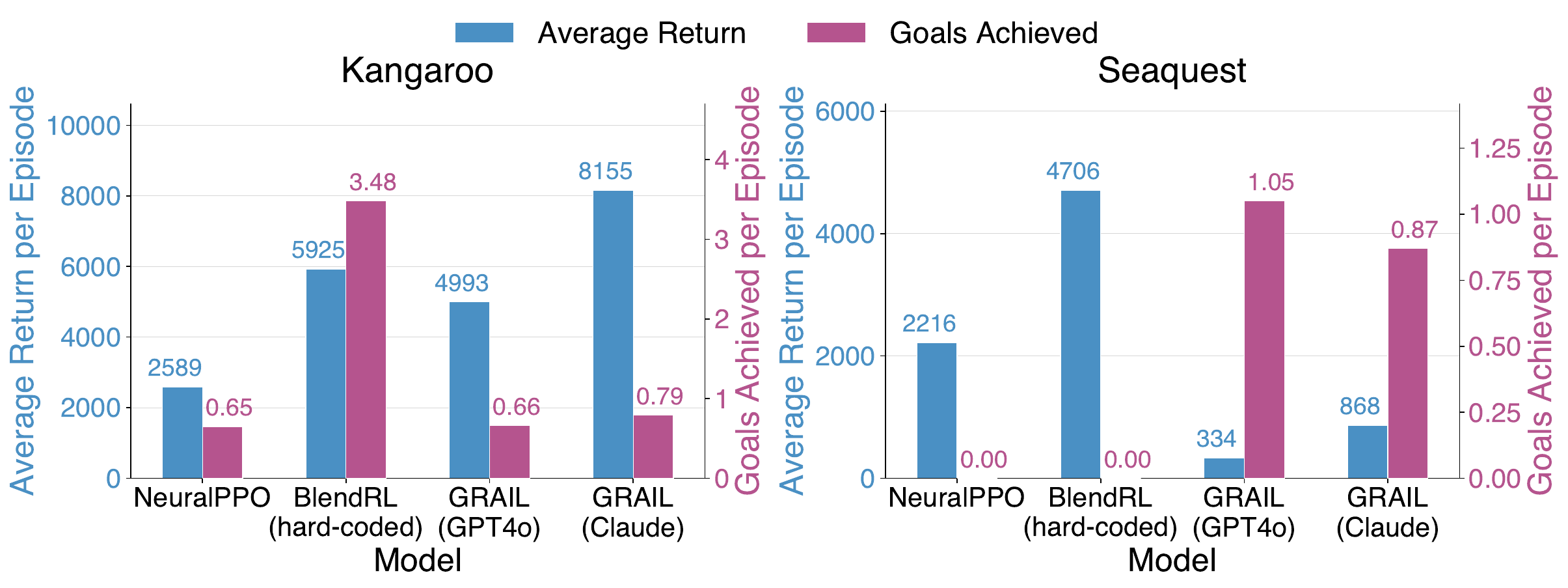}
    \caption{
        \textbf{Return vs.\ goal completion in the full environment (Stage~2).}
        Blue bars show average episodic return; pink bars show goals achieved per episode (reaching the child in Kangaroo; rescuing all divers in Seaquest).
        In Kangaroo, GRAIL (Claude) achieves the highest return but BlendRL's hand-coded strategy yields far more goals.
        In Seaquest, only GRAIL completes any goals, while baselines that maximize return fail to rescue divers entirely.
        Averages over 3 seeds (100 episodes); $c_{\OP{CA}} = 0.3$.
    }
    \label{fig:results_goals_2nd}
\end{figure}

To answer \textbf{Q3}, we evaluate each method in the complete environment, where both BlendRL and GRAIL jointly optimize their neuro-symbolic policies.
The spatial concepts established in the previous stages are kept fixed throughout this phase.
Figure~\ref{fig:results_goals_2nd} reports the average episodic return and goal completion for all baselines.

We distinguish two complementary success criteria: \emph{episodic return} (cumulative reward, including short-term gains such as defeating enemies) and \emph{goal completion} (achieving the environment’s high-level objective---reaching the child in Kangaroo or rescuing all divers in Seaquest).
These metrics can diverge, as an agent may maximize return through short-term actions without ever completing the long-horizon goal.

Our results reveal a consistent tension between these criteria.
In Kangaroo, GRAIL (Claude) achieves the highest return (8155), yet BlendRL’s hand-coded strategy yields far more goals (3.48 vs.\ 0.79 per episode), suggesting that expert-designed concepts are better tuned to this environment’s specific goal structure.
In Seaquest, the pattern reverses: BlendRL achieves the highest return (4706) but \emph{zero} goal completions, whereas GRAIL is the only method that successfully rescues divers (1.05 goals per episode for GPT-4o).
This demonstrates that high return does not imply meaningful task completion, and that GRAIL’s learned concepts enable qualitatively different behavior---pursuing high-level goals that reward-maximizing baselines neglect entirely.

This return-versus-goal tension highlights an open challenge in neuro-symbolic RL: jointly optimizing for reward and high-level goal completion.
In the following section, we analyze barriers to effective concept grounding and discuss potential paths forward.

\begin{figure}
    \centering
    \includegraphics[width=\linewidth]{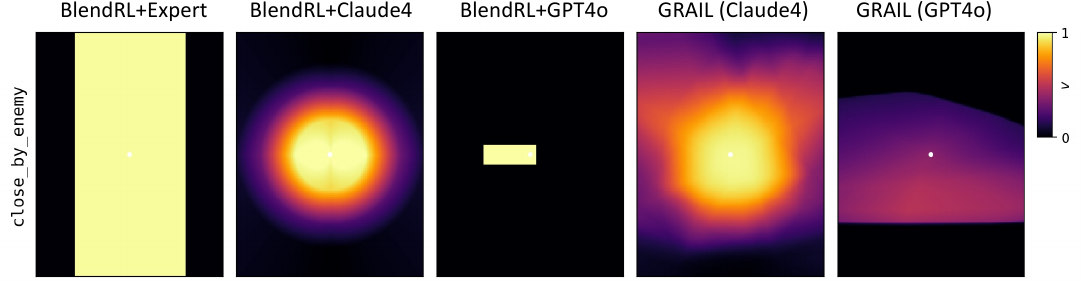}
    \caption{
        \textbf{Stage~2: learned $\mathtt{close\_by\_enemy}$ concept in Seaquest.}
        The blending module decides when to delegate control to the symbolic policy based on learned spatial predicates.
        Each heatmap shows the activation of $\mathtt{close\_by\_enemy}$ as a function of relative object position; white dots mark enemy positions.
        GRAIL produces broader, horizontally extended activations that capture lateral enemy movement, whereas hand-coded BlendRL functions yield near-uniform distributions and raw LLM proxies show high variability.
    }
    \label{fig:seaquest_2nd}
\end{figure}

Beyond aggregate performance, we examine whether the learned spatial concepts remain meaningful after joint training.
Figure~\ref{fig:seaquest_2nd} visualizes the spatial concepts acquired by the blending module in Seaquest.
The hand-coded BlendRL functions fail to capture the underlying environmental semantics, producing largely uniform distributions.
Raw proxy functions from Claude and GPT-4o yield inconsistent activation patterns that do not reliably reflect the spatial structure of the environment.
In contrast, GRAIL produces more coherent and adaptive representations: in both the Claude and GPT-4o settings, it broadens the activation map of $\mathtt{close\_by\_enemy}$, extending it horizontally to account for enemies that enter from both sides and move laterally---an adjustment well-aligned with the task’s demands.
These results demonstrate that GRAIL can adaptively ground spatial concepts even in complex environments with dynamic elements such as enemies.

\subsection{Ablation: Impact of Concept Alignment}
\label{sec:ablations}

\begin{figure}[!htbp]
    \centering
    \includegraphics[width=0.85\linewidth]{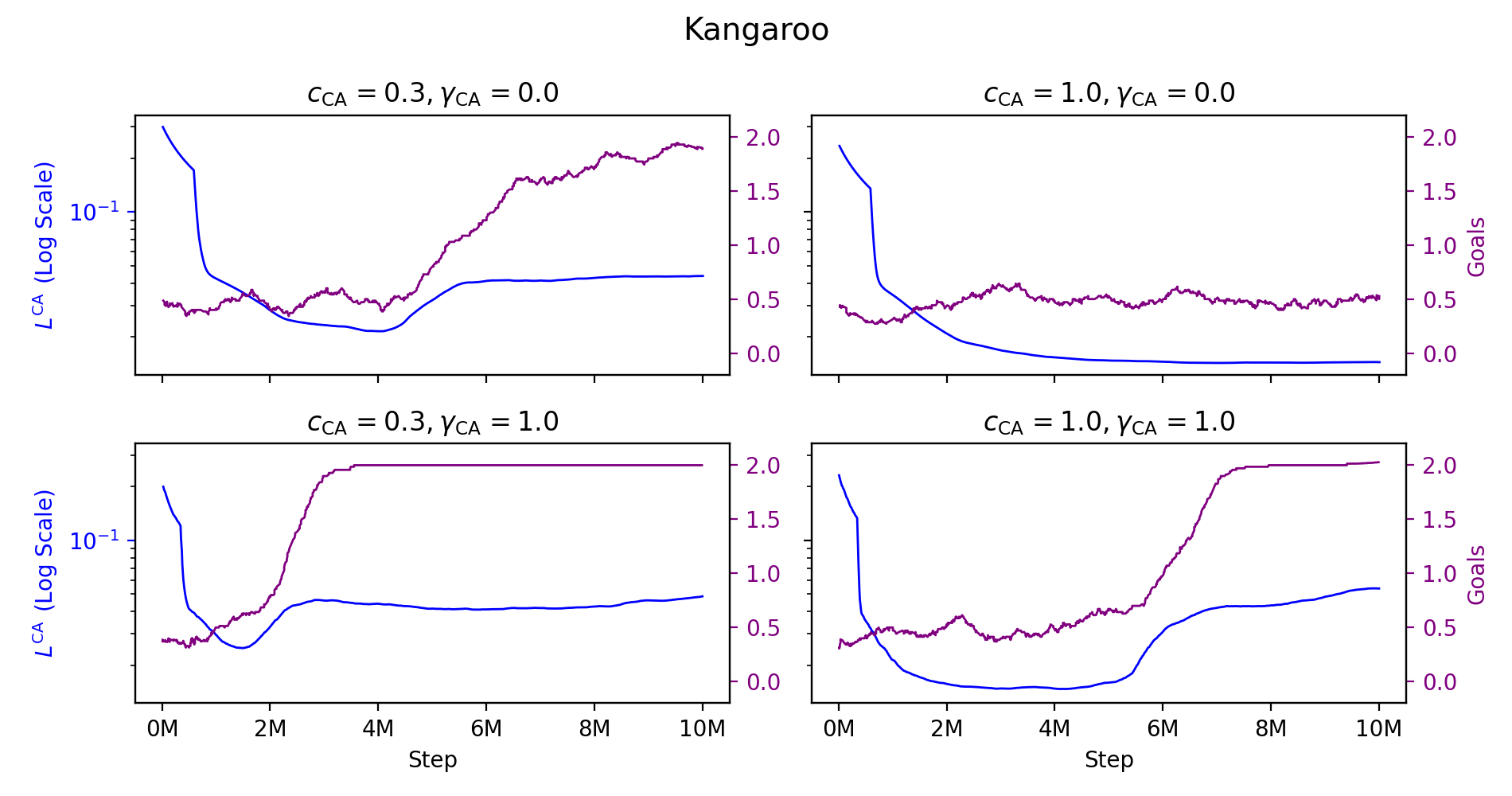}
    \caption{
        \textbf{Concept alignment loss $L^{\operatorname{CA}}$ vs.\ goals during Stage~1 training (Kangaroo).}
        Each panel shows a different combination of alignment coefficient $c_{\operatorname{CA}}$ and annealing factor $\gamma_{\operatorname{CA}}$.
        Performance improves as concepts diverge from the LLM proxies (rising $L^{\operatorname{CA}}$), and annealing ($\gamma_{\operatorname{CA}} = 1$) accelerates convergence.
    }
    \label{fig:ca_loss_vs_goals}
\end{figure}

Figure~\ref{fig:ca_loss_vs_goals} compares the concept alignment loss $L^{\operatorname{CA}}$ and the number of goals achieved during Stage~1 training in Kangaroo for different values of the alignment coefficient $c_{\operatorname{CA}} \in \{0.3, 1.0\}$ and the annealing factor $\gamma_{\operatorname{CA}} \in \{0, 1\}$.
Two observations stand out.
First, performance consistently improves as the learned concepts diverge from the LLM proxy functions---rising $L^{\operatorname{CA}}$ coincides with rising goals---indicating that the agent must move beyond the initial proxies to discover effective groundings.
Second, annealing the alignment loss ($\gamma_{\operatorname{CA}} = 1$) accelerates convergence and reduces sensitivity to the choice of $c_{\operatorname{CA}}$.
Together, these results suggest that strong initial guidance from the concept aligner, gradually attenuated over training, provides the best balance between alignment and adaptability.

The remaining hyperparameters are set as follows.
The grid resolution is $K = 49$, providing sufficient granularity to capture fine-grained spatial relationships while remaining computationally tractable ($L^{\operatorname{CA}}$ scales as $\mathcal{O}(K^2)$ per predicate per iteration).
The valuation function is a compact MLP ($2 \to 64 \to 32 \to 1$ with ReLU activations and a sigmoid output): the 2-dimensional input (relative offset) is low-dimensional, and a smaller network encourages smooth, interpretable groundings rather than overfitting to spurious patterns.

\subsection{Concept Misalignment Challenge}
To address \textbf{Q4}, we examine concept misalignment---cases where the agent learns spatial groundings that are systematically incorrect despite achieving reasonable returns.
Although GRAIL acquires useful concepts overall, our analysis reveals recurring failure patterns. We present a qualitative analysis of representative cases below.

\begin{figure}[!htbp]
    \centering
    \includegraphics[width=0.194\linewidth]{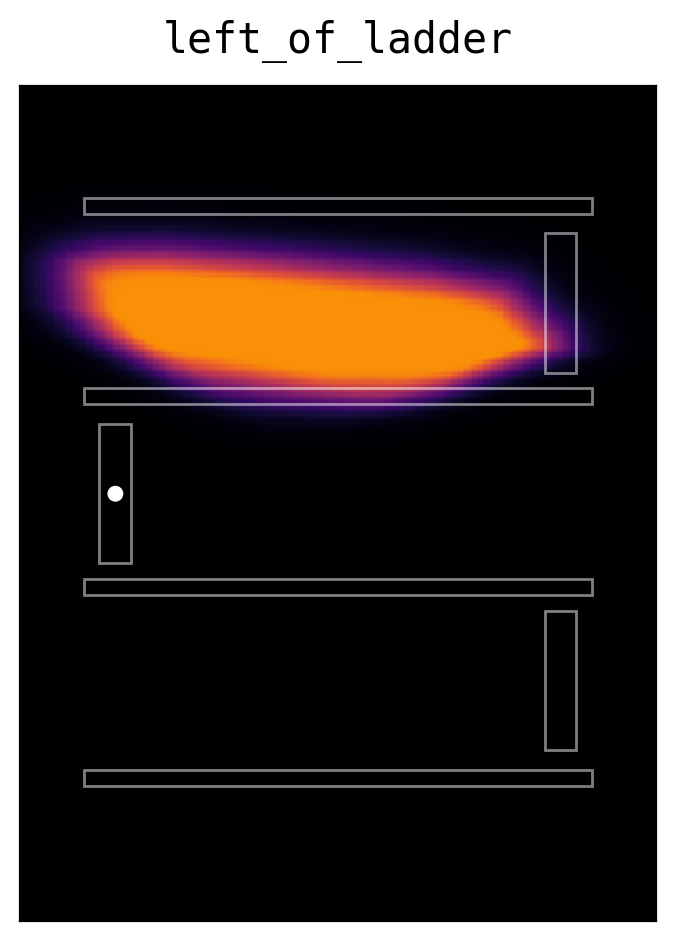}
    \includegraphics[width=0.194\linewidth]{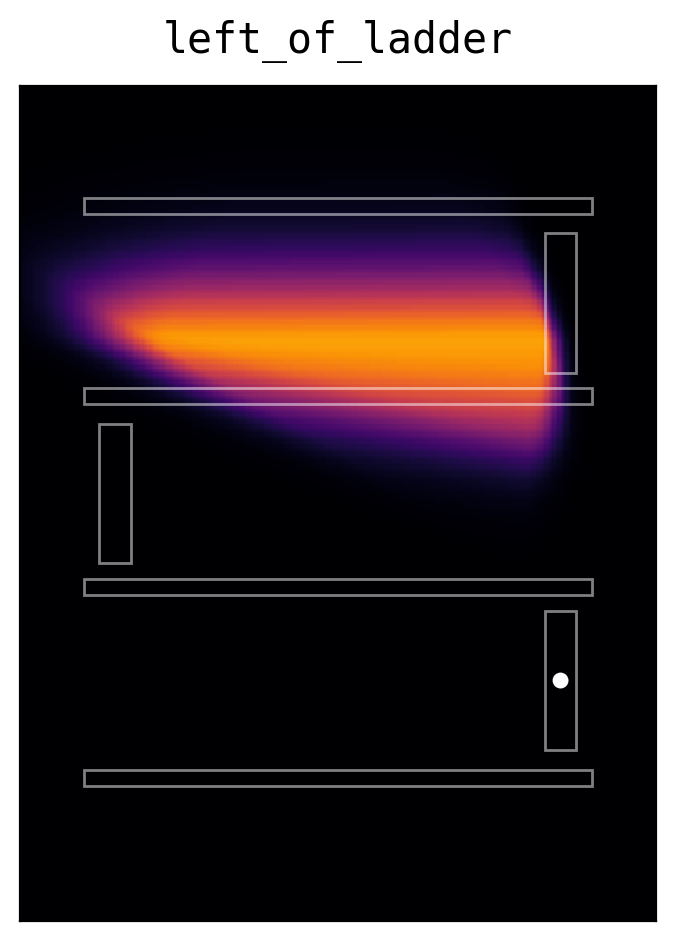}
    \includegraphics[width=0.194\linewidth]{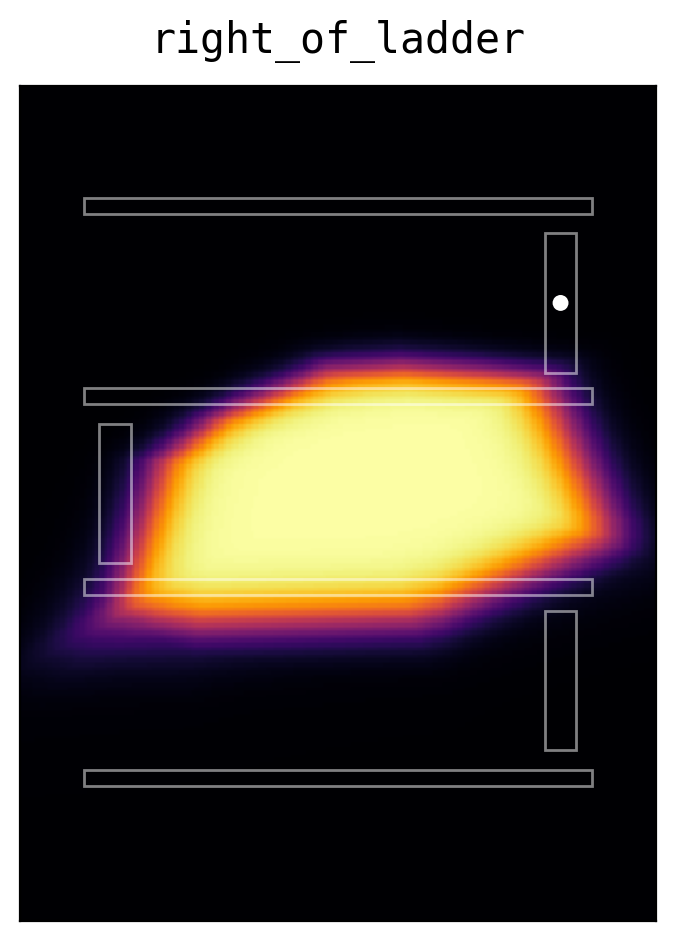}
    \includegraphics[width=0.194\linewidth]{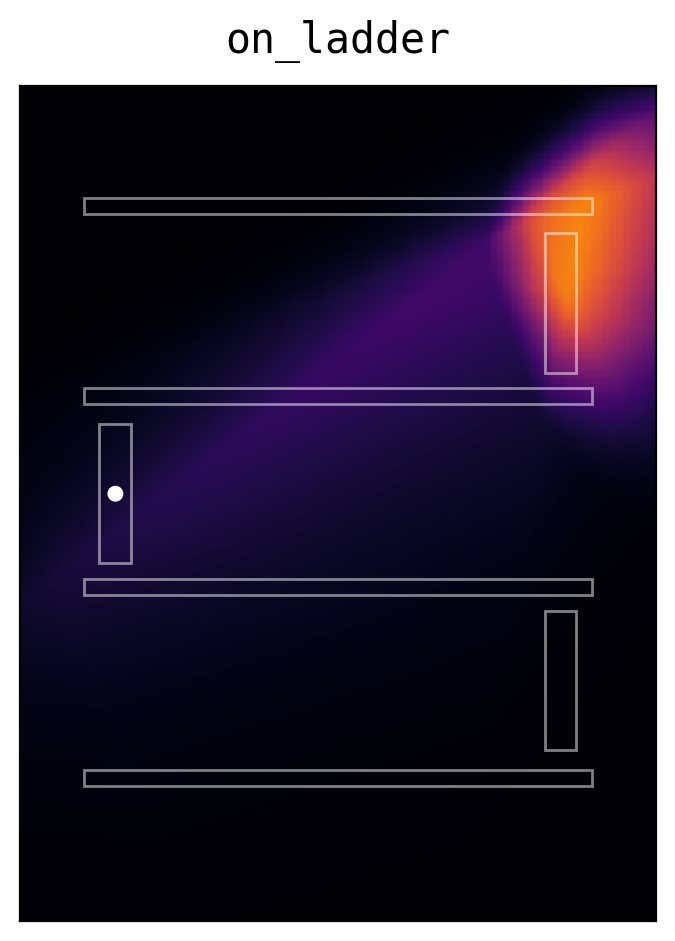}
    \includegraphics[width=0.194\linewidth]{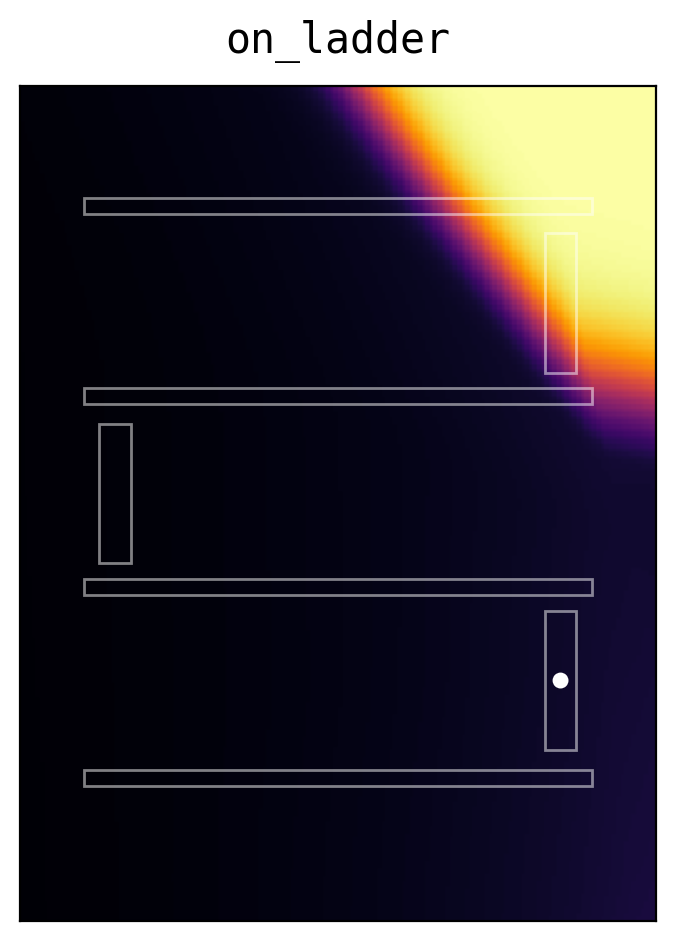}
    \caption{
        \textbf{Examples of misaligned spatial concepts in Kangaroo.}
        Each heatmap shows the learned truth value of a spatial predicate relative to a single ladder (white dot).
        Left two: $\mathtt{left\_of\_ladder}$ incorrectly activates for an unrelated ladder on a different platform.
        Center: $\mathtt{right\_of\_ladder}$ exhibits similar cross-platform confusion.
        Right two: $\mathtt{on\_ladder}$ is biased toward the top platform, likely due to the disproportionately high reward for reaching it.
        These misalignments typically arise when the alignment coefficient $c_{\operatorname{CA}}$ is set too low.
    }
    \label{fig:concept_misalignment_cc0}
\end{figure}

Figure~\ref{fig:concept_misalignment_cc0} illustrates misaligned spatial concepts in Kangaroo.
In the two leftmost examples, the agent incorrectly associates $\mathtt{left\_of\_ladder}$ with a ladder on a different platform.
A similar cross-platform confusion arises for $\mathtt{right\_of\_ladder}$ (center).
The two rightmost examples reveal a complementary failure mode: the agent's $\mathtt{on\_ladder}$ activation is biased toward the top platform even when evaluated relative to ladders on lower platforms, likely due to the disproportionately high reward for reaching the top.
These observations underscore that fully aligned concept acquisition remains a significant open challenge in neuro-symbolic reinforcement learning.
GRAIL mitigates this by introducing weak supervision at the predicate level, supplementing the action-level reward signal, but further work is needed to eliminate such systematic misalignments.

\section{Conclusion}
\label{sec:conclusion}

We introduced GRAIL, a neuro-symbolic reinforcement learning framework that acquires spatial concepts through direct interaction with the environment.
GRAIL leverages LLMs to generate proxy functions as weak supervision for concept grounding, and aligns these representations to each environment via a learnable concept aligner.
Across Kangaroo, Seaquest, and Skiing, GRAIL matched or exceeded strong neural and neuro-symbolic baselines, producing interpretable spatial concepts on par with hand-crafted valuation functions---without requiring expert-designed concept priors.

\paragraph{Scope and limitations.}
GRAIL automates the grounding of spatial predicate \emph{semantics}---the valuation functions that map object-pair offsets to truth values---while the predicate inventory, logic programs (Figure~\ref{fig:logic_programs}), and object-centric state extraction (OCAtari;~\citealp{OCAtari}) remain externally specified.
The two-stage training procedure, in which concepts are first learned in simplified environments (Stage~1) and then frozen during joint policy training (Stage~2), constitutes a form of curriculum learning analogous to~\citet{Mao_2019_NSCL}. This design prevents concept drift under short-term reward pressure but limits end-to-end adaptability.
The concept aligner serves as a warm-start rather than a hard constraint---performance improves as learned concepts \emph{diverge} from the proxies (Figure~\ref{fig:ca_loss_vs_goals})---yet a fundamentally incorrect proxy could still mislead early learning.
In our experiments, a single prompt template per environment and the first syntactically valid LLM output sufficed, suggesting reasonable robustness to LLM choice, though a systematic study of prompt sensitivity and corrupted proxies remains open.

\paragraph{Future work.}
A natural next step is end-to-end training that eliminates the two-stage split, allowing concepts to co-adapt with the full policy.
Extending GRAIL to $n$-ary and non-spatial predicates would broaden its applicability but requires richer input representations and proxy designs.
On the alignment side, replacing the current binary cross-entropy objective (Eq.~\eqref{eq:concept_alignment_loss}) with ranking or contrastive losses may improve robustness to proxy noise.
A particularly important direction is \emph{compositional} concept grounding, where coupled semantics across predicates---e.g., $\mathtt{between}(A,B,C)$ requiring joint reasoning over $\mathtt{left\_of}$ and $\mathtt{right\_of}$---are modeled through differentiable logical operators.
Further transparency could be gained by replacing MLPs with more interpretable architectures such as differentiable logic gate networks~\citep{Petersen_2022_DDLGN} or program synthesis~\citep{wust2024pix2code}.
Finally, scaling GRAIL to realistic embodied domains such as autonomous driving~\citep{li2022metadrive}, urban micromobility~\citep{wu2025metaurban}, and human-robot collaboration~\citep{puig2024habitat} would test its generality beyond Atari environments.

\section*{Acknowledgments}
This work was partly funded by the German Federal Ministry of Education and Research, the Hessian Ministry of Higher Education, Research, Science and the Arts (HMWK) within their joint support of the National Research Center for Applied Cybersecurity ATHENE, via the ``SenPai:XReLeaS'' project. The work has benefited from the Clusters of Excellence ``Reasonable AI'' (EXC-3057) and ``The Adaptive Mind'' (EXC-3066), both funded by the German Research Foundation (DFG) under Germany's Excellence Strategy.

\bibliographystyle{unsrtnat}
\bibliography{bibliography}
\end{document}